\documentclass[conference,compsoc]{IEEEtran}

\usepackage{tikz}
\usepackage{amsmath}
\usepackage{bbding}
\usepackage{booktabs}       
\usepackage{amsfonts}       
\usepackage{nicefrac}       
\usepackage{soul}
\usepackage{hyperref}
\usepackage{microtype}      
\usepackage{xcolor} 
\usepackage{epsfig}
\usepackage{graphicx}
\usepackage{amsmath}
\usepackage{amssymb}
\usepackage{multirow}
\usepackage{adjustbox}
\usepackage{color}
\usepackage{array}
\usepackage[dvipsnames]{xcolor}
\usepackage{enumitem}
\usepackage{multirow}
\usepackage{amsmath}
\usepackage{makecell}
\usepackage{hyperref}
\usepackage{ragged2e}
\usepackage{threeparttable}
\usepackage{relsize}
\usepackage{xspace}
\usepackage{algpseudocode}
\usepackage[normalem]{ulem}
\usepackage{bbm}
\usepackage[T1]{fontenc}
\usepackage[utf8]{inputenc}
\usepackage{tcolorbox}
\usepackage{tikz}
\usepackage{url} 
\usepackage{algpseudocode}
\usepackage{colortbl}
\usepackage{cellspace}
\usepackage{pifont}
\usepackage{tabularx}
\newcommand{\sys}{{\sc MixGuard}\xspace}
\renewcommand{\arraystretch}{0.9}
\usepackage[linesnumbered, ruled]{algorithm2e}
\usepackage[ruled,linesnumbered]{algorithm2e}
\usepackage{wasysym}
\usepackage{bbding}

\usepackage[table]{xcolor}

\sethlcolor{black}
\definecolor{mygray}{gray}{.9}
\definecolor{lightgray}{gray}{.9}
\definecolor{lightlightgray}{gray}{.95}

\newcommand{\halfsquare}{%
\begin{tikzpicture}[scale=0.18,baseline={(0,0)}]
  \fill (0,0) rectangle (0.5,1);
  \draw (0,0) rectangle (1,1);
\end{tikzpicture}%
}

\newcommand{\emptysquare}{%
\begin{tikzpicture}[scale=0.18,baseline={(0,0)}]
  \draw (0,0) rectangle (1,1);
\end{tikzpicture}%
}

\newcommand{\fullsquare}{%
\begin{tikzpicture}[scale=0.18,baseline={(0,0)}]
  \fill (0,0) rectangle (1,1);
  \draw (0,0) rectangle (1,1); 
\end{tikzpicture}%
}

\newcommand{\fulltriangle}{%
\begin{tikzpicture}[xscale=0.24,yscale=0.20,baseline={(0,0)}]
  \fill (0,0) -- (1,0) -- (0.5,0.866) -- cycle; 
  \draw (0,0) -- (1,0) -- (0.5,0.866) -- cycle;
\end{tikzpicture}%
}

\newcommand{\emptytriangle}{%
\begin{tikzpicture}[xscale=0.24,yscale=0.20,baseline={(0,0)}]
  \draw (0,0) -- (1,0) -- (0.5,0.866) -- cycle; 
\end{tikzpicture}%
}

\newcommand{\halftriangle}{%
\begin{tikzpicture}[xscale=0.24,yscale=0.20,baseline={(0,0)}]
  \draw (0,0) -- (1,0) -- (0.5,0.866) -- cycle;
  
  \begin{scope}
    \clip (0,0) -- (0.5,0) -- (0.5,0.866) -- (0,0.866) -- cycle;
    \fill (0,0) -- (1,0) -- (0.5,0.866) -- cycle;
  \end{scope}
\end{tikzpicture}%
}

\newcommand{\emptycircle}{%
\begin{tikzpicture}[scale=0.18,baseline={(0,0)}]
  \draw[line width=0.5pt] (0.5,0.5) circle (0.5);
\end{tikzpicture}%
}

\newcommand{\fullcircle}{%
\begin{tikzpicture}[scale=0.18,baseline={(0,0)}]
  \fill (0.5,0.5) circle (0.5);
  \draw[line width=0.5pt] (0.5,0.5) circle (0.5);
\end{tikzpicture}%
}
\newcommand{\halfcircle}{%
\begin{tikzpicture}[scale=0.18,baseline={(0,0)}]
  \begin{scope}
    \clip (0,0) rectangle (0.5,1);
    \fill (0.5,0.5) circle (0.5);
  \end{scope}
  \draw[line width=0.5pt] (0.5,0.5) circle (0.5);
\end{tikzpicture}%
}

\newcommand{\fulldiamond}{%
\begin{tikzpicture}[xscale=0.24,yscale=0.20,baseline={(0,0)}]
  \fill (0.5,0) -- (1,0.5) -- (0.5,1) -- (0,0.5) -- cycle; 
  \draw (0.5,0) -- (1,0.5) -- (0.5,1) -- (0,0.5) -- cycle;
\end{tikzpicture}%
}

\newcommand{\emptydiamond}{%
\begin{tikzpicture}[xscale=0.24,yscale=0.20,baseline={(0,0)}]
  \draw (0.5,0) -- (1,0.5) -- (0.5,1) -- (0,0.5) -- cycle; 
\end{tikzpicture}%
}

\newcommand{\halfdiamond}{%
\begin{tikzpicture}[xscale=0.24,yscale=0.20,baseline={(0,0)}]
  \fill (0.5,0) -- (0.5,1) -- (0,0.5) -- cycle;
  \draw (0.5,0) -- (1,0.5) -- (0.5,1) -- (0,0.5) -- cycle;
\end{tikzpicture}%
}

\newcommand{\halfpentagon}{%
\begin{tikzpicture}[xscale=0.22,yscale=0.20,baseline={(0,0)}]
  \begin{scope}
    \clip (0,0) rectangle (0.5,1);
    \fill (0.5,1) -- (0.976,0.655) -- (0.794,0.095)
          -- (0.206,0.095) -- (0.024,0.655) -- cycle;
  \end{scope}
  \draw (0.5,1) -- (0.976,0.655) -- (0.794,0.095)
        -- (0.206,0.095) -- (0.024,0.655) -- cycle;
\end{tikzpicture}%
}

\newcommand{\fullpentagon}{%
\begin{tikzpicture}[xscale=0.22,yscale=0.20,baseline={(0,0)}]
  \fill (0.5,1) -- (0.976,0.655) -- (0.794,0.095)
        -- (0.206,0.095) -- (0.024,0.655) -- cycle;
  \draw (0.5,1) -- (0.976,0.655) -- (0.794,0.095)
        -- (0.206,0.095) -- (0.024,0.655) -- cycle;
\end{tikzpicture}%
}

\makeatletter
\newcommand\blfootnote[1]{%
  \begingroup
  \renewcommand\thefootnote{}%
  \long\def\@makefntext##1{\parindent 1em\noindent ##1}%
  \footnotetext{#1}%
  \endgroup
}
\makeatother

\usepackage{tcolorbox}        
\tcbuselibrary{most}          
\tcbset{enhanced}             

\definecolor{lightred}{RGB}{254, 244, 236}
\definecolor{lightgreen}{RGB}{241, 248, 236}
\definecolor{lightblue}{RGB}{236, 243, 250}

\ifCLASSOPTIONcompsoc
  \usepackage[nocompress]{cite}
\else
  \usepackage{cite}
\fi
\ifCLASSINFOpdf
\else
\fi

\begin{document}

\title{The Art of Mixology: Mixup-based Obfuscation for Privacy-Preserving Split Learning in Large Language Models}

\author{
\IEEEauthorblockN{Chen Chen$^1$, Xiang Gao$^2$ Xianshun Wang$^2$, Chengran Li$^2$, Shengyu Xia$^2$, Xueluan Gong$^{1*}$, \\Linru Zhang$^1$, Qian Wang$^2$, and Kwok-Yan Lam$^1$}

$^1$College of Computing and Data Science, Nanyang Technological University, Singapore\\
$^2$School of Cyber Science and Engineering, Wuhan University, China\\
$^*$Corresponding author \\

\{chen.chen, xueluan.gong, linru.zhang, kwokyan.lam\}@ntu.edu.sg,\\
\{gaoxiang178, wangxianshun661, chengranli, shengyuxia, qianwang\}@whu.edu.cn
}

\maketitle
\begin{abstract}
Split learning provides a practical paradigm for resource-constrained users to train Large Language Models (LLMs) by offloading computation-intensive layers to a server while keeping raw data local. However, existing privacy-preserving split learning methods still face a difficult trade-off among utility, privacy, efficiency, and stability. Specifically, these methods often suffer from substantial utility degradation, remain vulnerable to advanced data reconstruction attacks, incur prohibitive computational and communication overhead, or exhibit unstable performance across different tasks. In this paper, we propose \sys, a novel mixup-based privacy-preserving split learning framework for LLMs. \sys introduces token-level obfuscation, representation-level obfuscation, and adaptive gradient perturbation mechanisms, which operate jointly to preserve useful learning signals while preventing privacy leakage to the server. Technically, \sys first constructs a lightweight calibration model on a public dataset to refine the approximated target representation, and then applies this model during privacy-preserving fine-tuning on private data. We conduct extensive experiments on four classification tasks and four text generation tasks across multiple LLM families, model sizes, architectures, and fine-tuning strategies. The results show that \sys preserves model utility comparable to non-split training baselines, consistently achieves stronger privacy protection than existing split learning defense methods against state-of-the-art data reconstruction attacks, and remains robust under adaptive attack settings. 
\end{abstract}


%
\IEEEpeerreviewmaketitle

\section{Introduction}
Large language models (LLMs) have advanced rapidly in recent years, demonstrating remarkable capabilities across a wide range of domains, including finance, healthcare, education, and scientific research \cite{zhao2023survey,bommasani2021opportunities}. These improvements are driven by scaling laws, where increasing model size and training data generally improve performance, generalization, and emergent capabilities \cite{hoffmann2022training}.  However, this scaling trend also substantially increases the cost of training, which is often beyond the capacity of resource-constrained edge clients, such as mobile devices, IoT devices, and smart sensors \cite{lin2024split}. To mitigate these challenges, distributed training paradigms for collaborative model development have been extensively explored \cite{mcmahan2017communication}. Among these approaches, Split Learning (SL) has emerged as a promising framework. \blfootnote{\emph{Mixology, a term from cocktail making, refers to blending drinks from multiple ingredients to reduce their individual harshness and bitterness.}}

Split learning (SL) partitions a model into multiple segments and distributes the corresponding computational workloads across different parties. A typical SL setting involves two parties: a client that owns the private data and a server that provides computational resources. This architecture aims to achieve two objectives: (1) preserving downstream fine-tuning utility and (2) protecting sensitive information exchanged during client-server communication. In vanilla linear SL, the client retains only the shallow layers, while the server hosts the remaining layers and loss computation \cite{thapa2022splitfed}. However, this configuration requires label disclosure. To mitigate label leakage, U-shaped SL is introduced to keep the lightweight head and tail layers on the client, while offloading the computation-intensive intermediate layers to the server \cite{lyu2023optimal, pham2023data}.  

\begin{table*}
	\centering
	\footnotesize
    \caption{A Comprehensive Comparison of Split Learning Methods. }
    \setlength\tabcolsep{3pt}
	\begin{tabular}{cccccccccccc}  
		\toprule
        \multirow{2}{*}{Method} & \multicolumn{2}{c}{Utility$^\dagger$} & \multicolumn{4}{c}{Privacy$^\mathsection$} & \multicolumn{2}{c}{Efficiency$^\star$} & Stability$^\ddagger$  \\
        \cmidrule(lr){2-3} \cmidrule(lr){4-7} \cmidrule(lr){8-9} \cmidrule(lr){10-10} 
        & Domain & Performance & Protected Data & Stage & Mechanism  & Robustness & Computation & Communication & Across Tasks \\
		\midrule 
        DP-Forward \cite{du2023dp} & Classification & \halfcircle & Input & Training & DP & \halfsquare & \fulldiamond & \fulltriangle  & \halfpentagon \\
        DP-SGD \cite{abadi2016deep} & Classification & \halfcircle & Labels & Training & DP & \emptysquare & \fulldiamond & \fulltriangle & \halfpentagon\\
        SAP \cite{shen2023split} & Classification & \emptycircle & Input & Training & DP & \halfsquare & \fulldiamond & \fulltriangle & - \\
        SnD \cite{mai2023split}  & Classification & \emptycircle & Input & Inference & DP &  \emptysquare & \halfdiamond & \halftriangle & \halfpentagon \\
        TPSL \cite{yang2022differentially} & Classification & \halfcircle & Labels & Training & DP & \halfsquare & \fulldiamond & \fulltriangle & -  \\
        KD-UFSL \cite{zaland2025guarding} & Classification & \halfcircle & Both & Training & DP & \halfsquare & \halfdiamond & \halftriangle & - \\
        SplitHE \cite{pereteanu2022split} & Classification & \halfcircle & Input & Inference & HE & \fullsquare & \emptydiamond & \emptytriangle & - \\
        PolyTransformer \cite{zimerman2024converting} & Classification & \halfcircle & Input & Inference & HE & \fullsquare & \emptydiamond & \emptytriangle & - \\
        HESplitNet \cite{nguyen2023split} & Classification & \halfcircle & Both & Training & HE & \fullsquare & \emptydiamond & \emptytriangle & - \\
        Split Ways \cite{khan2023split} & Classification & \halfcircle & Both & Training & HE & \fullsquare & \emptydiamond & \emptytriangle & - \\
        CURE \cite{kanpak2024cure} & Classification & \halfcircle & Both & Training & HE & \fullsquare & \halfdiamond & \halftriangle & - \\
        NoPeek \cite{vepakomma2020nopeek} & Classification & \halfcircle & Input & Training & RST & \halfsquare & \halfdiamond & \halftriangle & -  \\
        DualGuard \cite{liu2025dualguard} & Generation & \halfcircle & Both & Training & RST & \halfsquare & \halfdiamond & \halftriangle & - \\
        \sys  & Both & \fullcircle & Both & Training & Mixup & \fullsquare & \halfdiamond & \halftriangle & \fullpentagon \\
		\bottomrule
	\end{tabular}
	\label{tab:literature}
    \par\addvspace{4pt}
    \begin{tablenotes}[flushleft]
\begin{minipage}{\linewidth}
\footnotesize
\justifying
\setlength{\parindent}{0pt} 
\setlength{\itemsep}{0pt}
\setlength{\leftmargin}{0pt} 

\centering \item  {\footnotesize $\dagger$ \textbf{Utility.} \textit{Domain}: the original evaluated domains. \textit{Performance}: utility preservation compared to the non-private baselines. (\emptycircle\; = low; \halfcircle\; = comparable).}

\centering \item  {\footnotesize $\mathsection$ \textbf{Privacy.} \textit{Data}: the target data protected (e.g., input data, labels). \textit{Stage}: the protection aims to privatize training data or inference-time data. \textit{Mechanism}: underlying privacy-preserving techniques (e.g., DP, HE, or RST). \textit{Robustness}: ability to resist against data reconstruction attacks (\halfsquare\; = partial; \fullsquare\; = strong).}

\centering \item  {\footnotesize $\star$ \textbf{Efficiency.} \textit{Computation}: additional computational overhead compared to U-shape baseline (\emptydiamond\; = intensive cryptographic operations; \halfdiamond\; = lightweight model training; \fulldiamond\; = negligible extra overhead). \textit{Communication}: additional client-server communication cost (\emptytriangle\; = high; \halftriangle\; = moderate; \fulltriangle\; = no extra). 

\centering \item {\footnotesize $\ddagger$ \textbf{Stability.} \textit{Across Tasks}: the stability to preserve utility and privacy across tasks (\halfpentagon\; = stable on most tasks; \fullpentagon\; = stable across all tasks; - = not evaluated).}

}

\end{minipage}
 \end{tablenotes}
\end{table*}

Despite these architectural advances, existing privacy-preserving SL methods remain limited in several practical dimensions, as summarized in Table \ref{tab:literature}. \underline{First}, their utility has been evaluated primarily on classification tasks, leaving their applicability to more complex generation tasks underexplored. Some methods, particularly DP-based methods \cite{shen2023split,mai2023split} also incur noticeable utility degradation relative to non-private baselines. \underline{Second}, existing methods provide incomplete privacy coverage. Many of them protect only a single type of sensitive data, such as inputs or labels, or are designed for a specific phase, such as training or inference. More importantly, lightweight defenses, such as DP-SGD \cite{abadi2016deep} and SnD \cite{mai2023split}, often provide partial robustness against advanced data reconstruction attacks. \underline{Third}, stronger protection typically comes at the cost of efficiency. In particular, cryptographic methods introduce substantial computation and communication overhead \cite{pereteanu2022split, zimerman2024converting, nguyen2023split, khan2023split}, while other defenses may still require auxiliary-model operations \cite{vepakomma2020nopeek,liu2025dualguard}. \underline{Finally}, the effectiveness of existing methods is not consistently maintained across tasks \cite{du2023dp,abadi2016deep,mai2023split}. Their utility–privacy trade-offs can vary considerably across datasets, and their stability on challenging generative tasks remains insufficiently studied.

To address these challenges, we explore a new direction and propose \sys, a novel mixup-based privacy-preserving Split Learning framework for LLMs. \sys is inspired by mixup, a data-augmentation technique that combines inputs or hidden representations from multiple samples to improve training stability and generalization. We extend this principle to privacy protection through two complementary obfuscation mechanisms. First, \sys performs \textbf{token-level obfuscation} by sampling and inserting secret tokens into the input sequence. Second, the client applies \textbf{representation-level obfuscation} by jointly encoding the target sample with multiple auxiliary samples, linearly combining their hidden representations using secret vectors and matrices, and transmitting only the resulting mixed activations to the server. After trunk-side computation, the client decodes the returned messages to recover an approximation of the target representation. A lightweight calibration model further refines this approximation before it is passed to the tail model for loss computation. These two obfuscation mechanisms operate jointly to reduce the private forward signals exposed to the server. During backpropagation, \sys adaptively perturbs the gradients according to the calibration residual, which mitigates the backward leakage. Through this pipeline, \sys protects both inputs and labels in the U-shaped SL setting while preserving the utility of hidden representations and incurring limited computation and communication overhead.

We evaluate \sys on 4 classification tasks and 4 text generation tasks across two widely used LLM families. The results show that \sys preserves downstream fine-tuning utility compared with Non-split training baselines. We further assess its privacy protection against 4 state-of-the-art data reconstruction attacks, demonstrating that \sys consistently outperforms existing SL defense methods. Moreover, our experiments show that \sys remains effective across LLMs with diverse model sizes, architectures, and fine-tuning strategies. \sys also provides strong protection against adaptive attacks, which highlights its robustness in stronger adversarial settings.

Our work makes the following contributions:
\begin{itemize}[leftmargin=*]
\item We propose \sys, a novel mixup-based obfuscation mechanism for a privacy-preserving split learning framework with LLMs. To the best of our knowledge, \sys is the first mixup-based method designed for LLM split learning.
\item We introduce token-level obfuscation, representation-level obfuscation, and adaptive gradient perturbation modules, which operate jointly to preserve useful learning signals while preventing privacy leakage to the server.

\item We evaluate \sys across diverse tasks, LLM families,  and data reconstruction attacks. The results demonstrate that \sys preserves downstream fine-tuning utility, consistently outperforms existing defense baselines, and remains robust against adaptive attacks.

\end{itemize}

\section{Background}
\subsection{Large Language Model}
\label{sec:llm}
A Large Language Model (LLM) is a Transformer-based neural network that models a token sequence autoregressively \cite{vaswani2017attention}. Given an input sequence $x=(x_1,\ldots,x_T)$, the model estimates the
next-token distribution
\begin{equation}
p_{\theta}(x_t \mid x_{<t})
=
\operatorname{softmax}(\mathbf{W}\mathbf{h}_{t-1})_{x_t},
\end{equation}
where $\mathbf{H}^{(L)}=[\mathbf{h}_1,\ldots,\mathbf{h}_T]^{\top}
\in \mathbb{R}^{T\times d}$ denotes the final-layer hidden-state matrix,
and $\mathbf{h}_{t-1}\in\mathbb{R}^{d}$ is the hidden representation at
position $t-1$. $\mathbf{W}$ is the output projection.

The hidden-state matrices are produced by stacking $L$ Transformer blocks:
\begin{equation}
\mathbf{H}^{(\ell)}
=
\operatorname{Block}^{(\ell)}\left(\mathbf{H}^{(\ell-1)}\right),
\qquad \ell=1,\ldots,L.
\end{equation}
starting from token embeddings with positional encoding $\mathbf{H}^{(0)}=\mathrm{Embed}(x)+\mathrm{PosEnc}(x)$.
Each block mainly consists of self-attention and a position-wise feed-forward network. For self-attention, queries/keys/values are computed by linear projections of $\mathbf{H}$:
\begin{equation}
\mathbf{Q}=\mathbf{H}\mathbf{W}_Q,\quad \mathbf{K}=\mathbf{H}\mathbf{W}_K,\quad \mathbf{V}=\mathbf{H}\mathbf{W}_V,
\end{equation}
and the attention output is
\begin{equation}
\mathrm{Attn}(\mathbf{H})=\mathrm{softmax}\!\left(\frac{\mathbf{Q}\mathbf{K}^\top}{\sqrt{d_k}}+\mathbf{M}\right)\mathbf{V},
\end{equation}
where $\mathbf{M}$ is a mask (e.g., causal mask) to prevent attending to future tokens.

LLMs are commonly pretrained by minimizing the negative log-likelihood over large corpora:
\begin{equation}
\mathcal{L}_{\mathrm{LM}}(\theta)=-\sum_{t=1}^{T}\log p_\theta(x_t\mid x_{<t}),
\end{equation}
and later adapted via supervised fine-tuning on task-specific instruction-response data. The layered and modular design of Transformers also makes LLMs convenient to partition across devices in split learning, where intermediate activations can be transmitted at a chosen cut layer.

\subsection{Split Learning}
Split learning (SL) is a collaborative training paradigm in which a model is partitioned between a client (data owner) and a server (compute provider) \cite{gupta2018distributed}. Depending on the partition strategy, SL can be broadly categorized into label-sharing SL (L-Shape) and label-private SL (U-Shape). In L-Shape SL, the client processes private inputs through the shallow layers of the model and transmits the resulting intermediate activations (smashed data) to the server. The server completes the forward pass, computes the loss using the ground-truth labels, and propagates the gradients back to the client to update the client-side parameters. In the U-Shape SL, both the initial layers (head model) and the final layers (tail model) are retained on the client, while the intermediate layers (trunk model) are hosted on the server. The client first performs the head model forward pass, sends smashed data to the server for trunk model computation, and then receives the server-side output to complete the tail model forward pass and loss computation. During backpropagation, the gradients follow the reverse path to propagate. By keeping the labels on the client, this architecture mitigates label leakage \cite{kariyappa2023exploit}. 

Therefore, SL methods for LLMs commonly adopt the U-shaped architecture to provide stronger privacy protection.

\smallskip
\textbf{Data Reconstruction Attacks (DRA).} In SL, a key privacy requirement is that the sensitive training data should not be recovered from the information transmitted between the client and the server. However, this requirement is difficult to guarantee in practice. Recently, an increasing body of work has focused on DRAs, which demonstrate that smashed data in the forward pass and gradients in the backward pass may expose substantial information about private samples, which can be used to reconstruct inputs or infer labels \cite{chen2024unveiling, deng2021tag, balunovic2022lamp}. Specifically, SIP is a forward smashed-data inversion attack that reconstructs input texts from the transmitted intermediate activations by training an inversion model with auxiliary data and pre-trained client-side priors \cite{chen2024unveiling}. TAG and LAMP are backward gradient-matching attacks that exploit gradients returned during backpropagation to recover target sequences or label information \cite{deng2021tag, balunovic2022lamp}. BiSR further combines forward inversion and backward gradient matching into a bidirectional reconstruction strategy, producing stronger recovery than either direction alone \cite{chen2024unveiling}.

\smallskip
\textbf{Defense Mechanisms.}
To mitigate the risk of sensitive data leakage in SL, a variety of privacy-preserving defense methods have been proposed. Existing approaches can be broadly categorized into three families according to their underlying strategy, i.e., differential privacy (DP) \cite{du2023dp, abadi2016deep, mai2023split}, fully homomorphic encryption (FHE) \cite{pereteanu2022split, khan2023split}, and representation space transformations (RST) \cite{vepakomma2020nopeek, liu2025dualguard}.

\textit{DP-based approaches.} They protect SL by introducing noise into the information exchanged between the client and the server. In particular, DP-Forward perturbs the smashed data transmitted in the forward pass to reduce the private signals from intermediate representations \cite{du2023dp}. By contrast, DP-SGD operates on the backward pass by clipping per-sample gradients and adding noise during gradient updates \cite{abadi2016deep}. A key advantage of these methods is that they offer formal or semi-formal privacy guarantees. However, this protection typically comes at a substantial utility cost. Prior empirical studies have shown that the injected noise can severely distort training signals, leading to degraded convergence and reduced downstream performance \cite{pham2024enhancing}.

\textit{FHE-based approaches.} These methods aim to encrypt transmitted values while allowing specific computations to be performed on ciphertexts \cite{rho2024encryption}. In principle, this provides strong confidentiality guarantees, since the server has no access to the plaintext content. However, applying FHE to modern deep learning pipelines remains highly challenging because Transformer-based architectures rely heavily on non-linear operations, which are difficult and computationally expensive to approximate under current FHE schemes. Although recent studies have improved the feasibility of FHE-based inference and training \cite{zhang2024secure,moon2025thor,zhang2025moai}, they still incur substantial cryptographic computation and communication overhead. Consequently, their application to SL with LLMs is still impractical at scale.

\textit{RST-based approaches.} RST-based methods aim to reduce the invertibility or exploitable structure of smashed data by driving their distribution away from the input token representation. For example, NoPeek explicitly minimizes the statistical dependence between raw inputs and intermediate representations \cite{vepakomma2020nopeek}. DualGuard further reshapes the client-side representation space prior to fine-tuning to weaken reconstruction attacks against both inputs and labels \cite{liu2025dualguard}. Compared with FHE-based methods, RST-based defenses are often more lightweight and easier to integrate into practical SL pipelines. However, their privacy protection is typically heuristic rather than formally guaranteed, and they may also incur significant utility degradation.

\subsection{Mixup Techniques}
The mixup technique was originally introduced as a data augmentation strategy to improve model generalization by synthesizing training examples from pairs of samples \cite{zhang2017mixup}. In language models, existing mixup methods can be categorized into token-level mixup and representation-level mixup.

\smallskip
\textbf{Token-level Mixup.}
Token-level mixup constructs mixed textual inputs on the discrete tokens before they are mapped into continuous embeddings. Given two tokenized sequences $x_i$ and $x_j$, a token-level mixing operator can be written as
\begin{equation}
    \tilde{x}=\Phi(x_i, x_j),
\end{equation}
where each token $t \in \tilde{x}$ is selected from either $x_i$ or $x_j$ according to a predefined or stochastic mixing rule. The mixed sequence $\tilde{x}$ is then passed to the embedding layer and used for model training.

This technique has primarily been explored as a way to improve the generalization ability of language models. More recent work has extended token-level manipulation toward privacy protection \cite{feng2026mitigating}. 
For example, token obfuscation methods replace original tokens with semantically distinct but embedding-proximate shadow tokens. The intuition is that language models have some tokens that are close in embedding space while differing substantially in semantic meaning. Such substitutions can preserve representation-level similarity while weakening the semantic correspondence between the observed token sequence and the original input, thereby reducing the effectiveness of DRAs.

\smallskip
\textbf{Representation-level Mixup.}
This operation conducts linear interpolation between pairs of samples on their numerical inputs or hidden representations \cite{zhang2017mixup}. Given two training samples $(x_i, y_i)$ and $(x_j, y_j)$, the mixup constructs a synthesized sample by
\begin{equation}
    \tilde{x}=\lambda x_i + (1-\lambda)x_j,
\end{equation}
where $\lambda \in [0,1]$ is a mixing coefficient typically sampled from a Beta distribution. The training objective for model $f$ is defined as the corresponding interpolation of the two supervision signals:
\begin{equation}
\mathcal{L}_{\mathrm{mix}} = \lambda\,\mathcal{L}(f(\tilde{x}), y_i) + (1-\lambda)\,\mathcal{L}(f(\tilde{x}), y_j),
\end{equation}

While this strategy was first developed for continuous input domains such as image, extending it to natural language tasks is more challenging, where the discrete tokens cannot be interpolated as naturally as continuous pixel values. Therefore, existing methods perform mixup in hidden representation spaces \cite{sun2020mixup, chen2020mixtext} typically by interpolating the sample-level representations produced by the encoder $f_{1}$ before passing them to the classification head $f_{2}$. Formally, the mixed representation and loss are defined as 
\begin{align}
  \tilde{h} &= \lambda f_{1}(x_i) + (1-\lambda)f_{1}(x_j), \\
  \mathcal{L}_{\mathrm{mix}} &= \lambda\,\mathcal{L}\!\left(f_{2}(\tilde{h}), y_i\right) + (1-\lambda)\,\mathcal{L}\!\left(f_{2}(\tilde{h}), y_j\right).
\end{align}


Despite this progress, text-domain mixup is largely limited to classification tasks. Extending it to autoregressive generation is more challenging because contextual hidden states evolve across decoding steps, with each prediction conditioned on the preceding context. This sequential dependency makes direct representation interpolation problematic, which introduces substantial mutual semantic contamination between the mixed components. The resulting representation may distort the contextual signals required for accurate next-token prediction. Consequently, mixup remains less explored for text generation, particularly in LLM fine-tuning.

\section{Threat Model}
\textbf{Defender.} 
The defender corresponds to the client in the SL architecture. It owns a private training dataset, as well as public or synthetic datasets that can be used to train client-side auxiliary models. In addition, the defender has access to a support set, which may consist of arbitrary datasets or token sequences. The defender can generate private randomness, including random vectors, matrices, and tokens, that is not disclosed to the server. 

The defender does not assume prior knowledge of the server-side model architecture, nor of the specific data reconstruction strategies that the server may employ. During training, client-server communication is conducted through a secure channel, over which the client and server exchange smashed data in the forward pass and gradients in the backward pass. The defender's objective is to prevent reconstruction of its private training data from the smashed data and gradients, while maintaining model convergence and training accuracy comparable to non-private or centralized training. 

\textbf{Attacker.} 
We consider an honest-but-curious attacker on the server side in the SL architecture \cite{deng2021tag, balunovic2022lamp}. During training, the attacker can fully observe the exchanged smashed data and gradients. In addition, the attacker has access to the server-side model segment, architecture, and optimization procedure. However, the attacker has no direct access to the client's raw inputs and labels, or any client-side randomness (e.g., local random seed). The attacker may perform advanced reconstruction attacks based on the received smashed data and gradients. Additionally, the attacker may possess knowledge of the defender's mechanism and leverage adaptive data reconstruction attacks accordingly. The attacker's objective is to reconstruct sensitive information about the client's private training data. 

\section{Methodology}
\subsection{Overview of \sys}
\begin{figure*}[tt]
    \centering
    \includegraphics[width=0.97\textwidth]{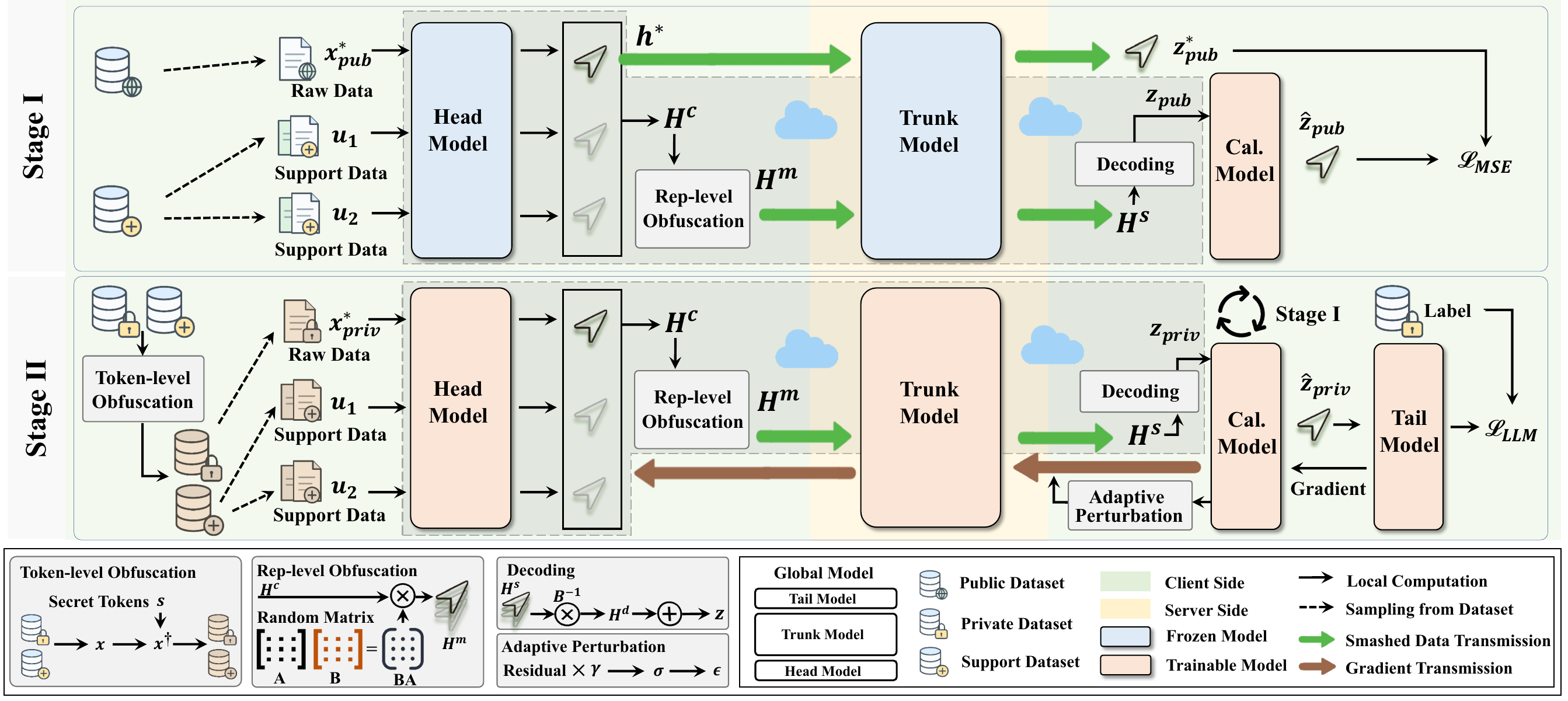}
    \caption{Overview of \sys. \sys employs a shared mixup-based split-learning protocol across two stages, as highlighted by the dotted boxes \raisebox{-0.25em}{\includegraphics[width=4.5ex, height=2.5ex]{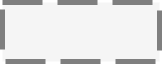}}. Stage I trains a lightweight calibration model using public data. Stage II performs privacy-preserving fine-tuning on private data. Rep and Cal. denote Representation and Calibration, respectively.} 
    \label{fig:overview}
\end{figure*}
\textbf{Intuition.} 
We formulate privacy preservation in SL as a controlled obfuscation problem. The objective is to obscure the sensitive client data from the server while keeping the training signal exploitable for the client. Existing approaches fail to fulfill both requirements. For example, DP-based methods introduce input-agnostic noise into smashed data. However, making discrete tokens indistinguishable requires large noise, which can push the smashed data away from the original representation space. This distribution shift often leads to systematic utility degradation.

To bridge this gap, we draw inspiration from Vicinal Risk Minimization theory \cite{chapelle2000vicinal} and mixup techniques \cite{zhang2017mixup}. We actively obfuscate private training samples by mixing multiple inputs and hidden representations to construct synthesized training instances. This technique was originally proposed for data augmentation, while we reinterpreted it in the context of privacy-preserving SL. A key advantage is that this approach often preserves useful training behavior even when the underlying model is highly nonlinear \cite{sun2020mixup, chen2020mixtext}. Based on this intuition, \sys introduces mixup-based obfuscation mechanisms to construct secret-augmented smashed data. Compared to DP noise, this strategy performs interpolation between existing representations and avoids significant distribution drift. At the same time, the mixture remains transparent to the client: the client retains full knowledge of the mixture composition and can disentangle it during subsequent local computation.

\smallskip
\textbf{Workflow.} \sys consists of two stages, i.e., calibration model construction and privacy-preserving fine-tuning (Figure \ref{fig:overview}). \underline{Stage I} constructs a lightweight calibration model using a public dataset. During this stage, only the calibration model is updated, while all global model segments (e.g., the head, trunk, and tail model) remain frozen. Specifically, the client performs the \textbf{representation-level mixup} by sampling secret coefficient vectors and matrices, and using them to encode multiple batches into mixed smashed data. The smashed data are transmitted to the server, processed by the frozen trunk, and returned to the client. The client then decodes the returned representations to obtain an approximated representation for the target batch, which serves as the input to the calibration model. In parallel, the client processes the corresponding target batch without mixup to obtain the ground-truth representation. The calibration model is trained to minimize the mean squared error (MSE) between the calibrated approximation and the ground truth. \underline{Stage II} performs the actual privacy-preserving SL fine-tuning on the private dataset. The client first applies \textbf{token-level obfuscation} by sampling secret tokens and inserting them into the original input. It then follows the same mixup-based procedure as in Stage I to obtain the target batch approximation. This approximated representation is fed into the calibration model to produce a refined activation, which is subsequently passed to the tail model for loss computation. To maintain the accuracy over the course of fine-tuning, the client periodically updates the calibration model using public data. During backpropagation, the client further introduces adaptive perturbation based on the calibration residual.

\subsection{Mixup-based SL Protocol}\label{sec:mixup-based-sl-protocol}

The two stages of \sys pursue different learning objectives, while they rely on the same underlying SL protocol. We therefore first formalize this common process, which provides the backbone for the procedures introduced in Sections \ref{sec:step-1} and \ref{sec:step-2}.  

We consider an $L$-layer global model $\mathcal{M}$ under the U-shape SL paradigms, where the model is partitioned into three contiguous segments:
\begin{equation}
\mathcal{M} = (\mathcal{M}_{\text{head}}, \mathcal{M}_{\text{trunk}}, \mathcal{M}_{\text{tail}}), 
\end{equation}
with cut layers $l_{1}$ and $l_{2}$ ($0 \leq l_{1} < l_{2} \leq L$). The client hosts $(\mathcal{M}_{\text{head}}, \mathcal{M}_{\text{tail}})$, while the server hosts $\mathcal{M}_{\text{trunk}}$.  Notably,  $\mathcal{M}_{\text{tail}}$ comprises not only the final several layers of $\mathcal{M}$, but also the model head responsible for token classification. To simplify notation, we denote the corresponding forward mappings of $\mathcal{M}_{\text{head}}$, $\mathcal{M}_{\text{trunk}}$, and $\mathcal{M}_{\text{tail}}$ by $g_1$, $g_2$, and $g_3$, respectively. Accordingly, the overall forward mapping of $\mathcal{M}$ is given by:

\begin{equation}
    g = g_3 \circ g_2 \circ g_1
\end{equation}

The client further has access to an auxiliary non-private public (or synthesized) dataset $\mathcal{D}_{\text{pub}} = (\mathcal{X}_{\text{pub}}, \mathcal{Y}_{\text{pub}})$ and a private dataset $\mathcal{D}_{\text{priv}} = (\mathcal{X}_{\text{priv}}, \mathcal{Y}_{\text{priv}})$, where $\mathcal{X}$ and $\mathcal{Y}$ represent the input and output set, respectively. In addition, the client maintains a secret support set $\mathcal{D}_{\text{supp}} = \mathcal{X}_{\text{supp}}$, which may consist of arbitrary text samples or token sequences unavailable to the attacker.

\smallskip
\noindent\textbf{Head Model Encoding.}
For each training step, the client samples a target input text $x^{*} \in \mathcal{X}$

. The client additionally samples $k-1$ support data $\{u_1,\ldots,u_{k-1}\} \subseteq \mathcal{X}_{\text{supp}}$. These samples form a size-$k$ input set
\begin{equation}
X = \{x^{*}, u_1,\ldots,u_{k-1}\}, \qquad |X| = k.
\end{equation}
For each $x_i \in X$, the client tokenizes the text and maps the resulting token sequences into an embedding matrix $\mathbf{e}_i \in \mathbb{R}^{n_i \times d}$, where $n_i$ denotes the sequence length and $d$ is the embedding dimension. To enable embedding-wise mixing across samples, the client pads the embedding matrix to a common length $n = \max(\{n_i\}_{i=1}^{k})$, yielding padded embedding $\Tilde{\mathbf{e}}_i \in \mathbb{R}^{n \times d}$. The padded embedding is then encoded by the $\mathcal{M}_{\text{head}}$, producing hidden representations $\mathbf{h}_i = g_1(\Tilde{\mathbf{e}}_i) \in \mathbb{R}^{n \times d}$. The client processes all $x_i \in X$, and stacks the resulting $k$ hidden representations along a new leading dimension to construct a tensor $\mathbf{H}^c = \text{stack}(\{\mathbf{h}_i\}_{i=1}^{k}) \in \mathbb{R}^{k \times n \times d}$, where $\text{stack}(\cdot)$ denotes the stacking operation. Since the target representation $\mathbf{h}^{*}$ corresponding to the input text $x^{*}$ appears at a random but client-tracked position within $\mathbf{H}^c$, we denote its index as $i^{*}$.

\smallskip
\noindent\textbf{Representation-level Obfuscation.} The stacked client-side representations in $\mathbf{H}^{c}$ contain substantial plaintext information that requires protection. To address this issue, we introduce a representation-level obfuscation mechanism. Specifically, \sys conceals $\mathbf{h}^{*}$ among $k-1$ representations from a secret support set through a secret Sign Vector $v \in \mathbb{R}^m$ and two secret matrices, i.e., a Mixing Matrix $\mathbf{A}\in\mathbb{R}^{m\times k}$ and an invertible Blinding Matrix $\mathbf{B}\in\mathbb{R}^{m \times m}$.

\smallskip
\noindent\emph{Sign Vector.} The sign vector $\mathbf{v}$ is a secret vector with entries $v_i$ independently sampled from $\{-1, +1\}$.

\smallskip
\noindent\emph{Mixing Matrix.}

The client randomly samples a positive-value mixing matrix $\mathbf{A}=[a_{i,j}] \in \mathbb{R}^{m\times k}$ ($a_{i, j} > 0$), which linearly combine the $h$ hidden representations in $\mathbf{H}^c$ into $m$ mixed messages. To ensure that the client can later recover the signal corresponding to $x^{*}$, the matrix $\mathbf{A}$ is required to satisfy the following constraint:

\begin{itemize}[leftmargin=*]
\item Column-wise Sum Constraint. This constraint requires that summing the $m$ mixed output messages cancels all non-target representations while preserving unit weight on the target representation. Formally,
\begin{align}
\sum_{i=1}^{m} v_{i}a_{i,j} &=
\begin{cases}
1, & j = i^*, \\
0, & \forall j\in\{1, 2,\ldots,k\}\textbackslash\{i^*\}.
\end{cases} \label{eq:colsum-sum-constraint}
\end{align}
\item Row-wise Support Constraint. This constraint captures the minimal requirements of the mixup-based privacy preservation, where each transmitted message must involve at least two source representations. This can be written as
\begin{align}
\|\mathbf{A}_{i,:}\|_0 &\ge 2,\quad \forall i\in\{1,\ldots,m\}, \label{eq:row_nnz}
\end{align}
where $\mathbf{A}_{i,:}$ denotes the $i$-th row of $\mathbf{A}$ and $\|\cdot\|_0$ counts the number of non-zero entries. 
\end{itemize}

\smallskip
\noindent\emph{Blinding Matrix.}
Mixing Matrix $\mathbf{A}$ alone is insufficient to fully obfuscate the private signals. For example, a server-side attacker may aggregate the received messages $\sum_{j=1}^{k}\big(\sum_{i=1}^{m}a_{i,j}\,\mathbf{h}_j\big)=\sum_{i=1}^{m}\big(\sum_{j=1}^{k}a_{i,j}\,\mathbf{h}_j\big) = \mathbf{h}_{i^{*}} = \mathbf{h}^{*}$, which may directly reveal the target representation $\mathbf{h}^{*}$.
To further obfuscate the private signals, the client applies an additional invertible Blinding Matrix $\mathbf{B}$, which further obfuscates the $m$ mixed messages before transmission. Accordingly, the resulting smashed data from the client is 
\begin{align}
\mathbf{H}^m &\;=\; \mathbf{H}^c\times_{1} \mathbf{A} \times_{1} \mathbf{B}, \\
&\;=\; \mathbf{H}^c\times_{1} (\mathbf{B}\mathbf{A}), 
\end{align}
where $\times_1$ denotes the mode-1 tensor-matrix product\footnote{Mode-1 product  

$(\mathbf{H} \times_1 \mathbf{A})_{i,:,:} \;=\; \sum_{j=1}^{k} \mathbf{A}_{i,j}\, \mathbf{H}_{j,:,:},\quad i=1,..,m,$
}. Consequently, the obfuscated tensor $\mathbf{H}^m \in \mathbb{R}^{m \times n \times d}$ is the smashed data actually transmitted to the server. We denote each representation in this tensor by $q_i$, i.e., $q_i = \mathbf{H}^{m}_{i,:,:}, i \in \{1,\dots, m\}$. Notably, both $\mathbf{A}$ and $\mathbf{B}$ are sampled independently for each training batch.

\smallskip
\noindent\textbf{Trunk Model Encoding.}
Upon receiving the smashed tensor $\mathbf{H}^m$, the server applies the forward mapping $g_2(\cdot)$ of the computation-intensive trunk model $\mathcal{M}_{\text{trunk}}$ to each slice along the first mode, and produce the server-side tensor $\mathbf{H}^s \in \mathbb{R}^{m \times n \times d}$. Specifically, for $i \in \{1, 2, \dots, m\}$, $\mathbf{H}^s_{i,:,:} = g_2(\mathbf{H}^m_{i,:,:})$. This server-side smashed tensor $\mathbf{H}^s$ is then returned to the client, according to the U-shape SL paradigm. 

\smallskip
\noindent\textbf{Tail Model Decoding.}
The server-side tensor $\mathbf{H}^s$ contains the encoded outputs from the mixed input batches under the mixing and blinding matrices $\mathbf{A}$ and $\mathbf{B}$. To estimate the target-related signal, the client first reverses the blinding effect by applying $\mathbf{B}^{-1}$: 
\begin{equation}
\mathbf{H}^{d} \;=\; \mathbf{H}^s\times_{1} \mathbf{B^{-1}}.
\end{equation}
The client then constructs an approximation of the target representation of $x^{*}$ at layer $l_s$ by summing the representations of $\mathbf{H}^s$ along the first mode. Specifically,
\begin{equation}
\mathbf{z} \;=\; \sum_{i=1}^{m} v_i\mathbf{H}^{d}_{i,:,:}
\;\in\; \mathbb{R}^{n \times d}.\label{eq:approximated-representation}
\end{equation}
This operation follows the column-wise sum constraint defined in Eq.\ref{eq:colsum-sum-constraint} on $\mathbf{H}^d$.

Within this mixup-based SL protocol, the raw input data $x^*$ never leaves the client. The server observes only obfuscated hidden representations without access to (i) the support samples involved in the obfuscation process or (ii) the secret mixing and blinding coefficients $(\mathbf{A}, \mathbf{B})$. 
 
The protocol defines a transformation from a raw input $x^{*}$ to its approximated representation $\mathbf{z}$, which we denote by $F: \mathcal{X} \rightarrow \mathbb{R}^{n \times d}$. This procedure is shared by both stages of \sys, i.e., calibration model construction and privacy-preserving fine-tuning. The mixup-based SL protocol is summarized in Algorithm \ref{alg:mixup_sl_protocol}.

\subsection{Stage I: Calibration Model Construction}\label{sec:step-1}

\sys framework relies on the approximated representation $\mathbf{z}$ in Eq.\ref{eq:approximated-representation} for SL with strong utility retention. However, despite preserving useful training signals, $\mathbf{z}$ generally deviates from the true $l_{s}$-layer representation $\mathbf{z}^{*}$, the ground-truth trunk output of $\mathbf{h}^{*}$. This discrepancy is primarily caused by the non-linear transformations within LLMs and may degrade downstream optimization. To mitigate this issue, we introduce a lightweight calibration model $\mathcal{M}_{\text{cal}}$ that learns to correct the approximation error, and transforms the approximated representation $\mathbf{z}$ into a refined representation $\hat{\mathbf{z}}$ better aligned with $\mathbf{z}^{*}$. The corresponding forward mapping is denoted as $g_{cal}: \mathbb{R}^{n \times d} \rightarrow \mathbb{R}^{n \times d}$. This model is implemented as a residual network with multiple calibration blocks sequentially stacked. Given an input representation $\mathbf{z} \in \mathbb{R}^{n \times d}$, each calibration block learns a low-rank residual transformation $\Delta (\mathbf{z})$ through a bottleneck architecture with GELU function \cite{hendrycks2016gaussian} and dropout \cite{hinton2012improving}.
\begin{align}
\operatorname{Block}(\mathbf{z}) &= \mathbf{z} + \Delta (\mathbf{z})\\
&=\mathbf{z}+\left(\operatorname{Dropout}\!\left(\operatorname{GELU}\!\left(\mathbf{z}\; W_{\mathrm{down}}\right)\right)\right)W_{\mathrm{up}},
\end{align}
where $W_{\text{down}} \in \mathbb{R}^{d \times r}$ projects the hidden representation into a low-dimensional bottleneck space, $W_{\text{up}} \in \mathbb{R}^{r \times d}$ maps it back to the original hidden space, and $r$ denotes the bottleneck dimension. We use $r = 64$ by default.

\smallskip
\noindent\textbf{Calibration Model Training.}
We pre-train the calibration model $\mathcal{M}_{\text{cal}}$ using the public (or synthesized) dataset $\mathcal{X}_{\text{pub}}$. For each target sample $x^{*}_{\text{pub}} \in \mathcal{X_{\text{pub}}}$, the client and server first construct its approximated $l_{s}$-layer representation $\mathbf{z}_{\text{pub}} \in \mathbb{R}^{n \times d}$ by following the mixup-based SL protocol, i.e., $\mathbf{z}_{\text{pub}} = F(x^{*}_{\text{pub}})$. The corresponding target representation $\mathbf{z}^{*}_{\text{pub}} \in \mathbb{R}^{n \times d}$ is then obtained through a standard forward pass without mixup, i.e., transmitting the plain smashed data $\mathbf{h}^{*}$, which is further encoded by the trunk model. This process can be expressed as 
\begin{equation}
    \mathbf{z}^{*}_{\text{pub}} \;=\; (g_{2} \circ g_{1})(x^{*}_{\text{pub}}).
\end{equation}
Since $x^{*}_{\text{pub}}$ is sampled from $\mathcal{X}_{\text{pub}}$, this communication does not expose any private data.
\newcommand{\CommentStyle}[1]{\scriptsize\ttfamily #1}
\SetCommentSty{CommentStyle}
\begin{algorithm}[t]
\scriptsize
\caption{Mixup-based SL Protocol}
\label{alg:mixup_sl_protocol}
\KwIn{
Target sample $x^{*}$, support set $\mathcal{X}_{\text{supp}}$, 
client head $g_1$, server trunk $g_2$, 
source size $k$, mixture size $m$
}
\KwOut{
Approximated target representation $\hat{\mathbf{z}}$
}

\SetKwFunction{MixupSL}{MixupSL}
\SetKwProg{Fn}{Function}{:}{}
\Fn{\MixupSL{$x^{*}, \mathcal{X}_{\text{supp}}, g_1, g_2, k, m$}}{

Sample $k-1$ support samples from $\mathcal{X}_{\mathrm{supp}}$ and form 
$X=\{x^{*},s_1,\ldots,s_{k-1}\}$\;

\ForEach{$x_i\in X$}{
    $\mathbf{h}_i \leftarrow g_1(x_i)$ \tcp*[r]{client representations}
}
$\mathbf{H}^{c}\leftarrow \mathrm{stack}(\{\mathbf{h}_i\}_{i=1}^{k})$\;

Sample a secret sign vector $\mathbf{v}\in\{-1,+1\}^{m}$, a mixing matrix $\mathbf{A}$ and an invertible blinding matrix $\mathbf{B}$\;

$\mathbf{H}^{m}\leftarrow \mathbf{H}^{c}\times_1(\mathbf{B}\mathbf{A})$ \tcp*[r]{obfuscation}

Transmit $\mathbf{H}^{m}$ to the server;

$\mathbf{H}^{s}_{i,:,:}\leftarrow g_2(\mathbf{H}^{m}_{i,:,:})$ \tcp*[r]{trunk representations}

Transmit $\mathbf{H}^{s}$ back to the client;

$\mathbf{H}^{d}\leftarrow \mathbf{H}^{s}\times_1 \mathbf{B}^{-1}$ \tcp*[r]{remove the blinding effect}

$\hat{\mathbf{z}}\leftarrow \sum_{i=1}^{m}v_i\mathbf{H}^{d}_{i,:,:}$ \tcp*[r]{aggregate representations}

\Return{$\hat{\mathbf{z}}$}\;
}
\end{algorithm}
Given paired supervision $(\mathbf{z}_{\text{pub}}, \mathbf{z}^{*}_{\text{pub}})$, we train $\mathcal{M}_{\text{cal}}$ using the Mean Squared Error (MSE) objective: 
\begin{equation}
\mathcal{L}_{\mathrm{MSE}}(\mathbf{z}_{\text{pub}}, \mathbf{z}^{*}_{\text{pub}})
= \frac{1}{nd}\left\lVert g_{\text{cal}}(\mathbf{z}_{\text{pub}}) - \mathbf{z}^{*}_{\text{pub}} \right\rVert^2,
\end{equation}
where $\mathbf{z}_{\text{pub}}, \mathbf{z}^{*}_{\text{pub}} \in \mathbb{R}^{n \times d}$, and $\left\|\cdot\right\|$ denotes $\ell_2$ norm. During this stage, only the parameters of $\mathcal{M}_{\text{cal}}$ are updated, while $\mathcal{M}_{\text{head}}$ and $\mathcal{M}_{\text{trunk}}$ remain frozen. Algorithm \ref{alg:calibration_construction} formalizes the calibration model construction procedure.
\begin{algorithm}[t]
\scriptsize
\caption{Calibration Model Construction}
\label{alg:calibration_construction}
\KwIn{
Public dataset $\mathcal{D}_{\text{pub}}=(\mathcal{X}_{\text{pub}},\mathcal{Y}_{\text{pub}})$,
support set $\mathcal{X}_{\text{supp}}$,
head model $g_1$, trunk model $g_2$,
calibration model $g_{\text{cal}}$,
source size $k$, mixture size $m$
}
\KwOut{
Trained calibration model $g_{\text{cal}}$
}

\SetKwFunction{MixupSL}{MixupSL}
\SetKwProg{Fn}{Function}{:}{}

\Fn{\textsc{CalibrationTraining}$(\mathcal{D}_{\text{pub}}, \mathcal{X}_{\text{supp}}, g_1, g_2, g_{\text{cal}}, k, m)$}{

Freeze $g_1$ and $g_2$\;

\ForEach{$x^{*}_{\text{pub}}\in \mathcal{X}_{\text{pub}}$}{

    $\hat{\mathbf{z}}_{\text{pub}} \leftarrow$ 
    \MixupSL{$x^{*}_{\text{pub}}$, $\mathcal{X}_{\text{supp}}$, $g_1$, $g_2$, $k$, $m$}

    $\mathbf{h}^{*}_{\text{pub}}\leftarrow g_1(x^{*}_{\text{pub}})$\;

    $\mathbf{z}^{*}_{\text{pub}}\leftarrow g_2(\mathbf{h}^{*}_{\text{pub}})$
    \tcp*[r]{true representation}

    $\mathbf{z}_{\text{pub}}\leftarrow g_{\text{cal}}(\hat{\mathbf{z}}_{\text{pub}})$\;

    $\mathcal{L}_{\text{MSE}} \leftarrow 
    \frac{1}{nd}\|\mathbf{z}_{\text{pub}}-\mathbf{z}^{*}_{\text{pub}}\|_2^2$\;

    Update $g_{\text{cal}}$ by minimizing $\mathcal{L}_{\text{MSE}}$\;
}

\Return{$g_{\text{cal}}$}\;
}
\end{algorithm}

\subsection{Stage II: Privacy-Preserving Fine-tuning}\label{sec:step-2}
This stage performs the actual fine-tuning on the private dataset $\mathcal{D}_{\text{priv}}$, while protecting it from leakage. The calibration model $\mathcal{M}_{\text{cal}}$ trained in Stage I is employed to obtain a more accurate representation for optimizing the global model parameters, i.e., $\mathcal{M} = (\mathcal{M}_{\text{head}}, \mathcal{M}_{\text{trunk}}, \mathcal{M}_{\text{tail}})$.  

\smallskip
\noindent\textbf{Token-level Obfuscation.} 
We first sample secret random tokens $s$ from the vocabulary and insert them into the raw private inputs. Formally, for each sample $x \in \mathcal{X}_{\text{priv}} \cup \mathcal{X}_{\text{supp}}$, the client applies this obfuscation to produce a token sequence:
\begin{equation}
    x^{\dagger} = \Phi(x, s)
\end{equation}
This process yields the obfuscated private and support datasets $\mathcal{X}^{\dagger}_{\text{priv}}$ and $\mathcal{X}^{\dagger}_{\text{supp}}$. 


\smallskip
\noindent\textbf{Representation-level Obfuscation.}
Following the mixup-based SL protocol, the approximation $\mathbf{z}_{\text{priv}} = F(x^{*}_{\text{priv}})$ is constructed from the private input $x^{*}_{\text{priv}} \in \mathcal{X}^{\dagger}_{\text{priv}}$ and $\mathcal{X}^{\dagger}_{\text{supp}}$. Using the calibration model $\mathcal{M}_{\text{cal}}$, the client produces a refined representation:
\begin{equation}
\hat{\mathbf{z}}_{\text{priv}} = g_{\text{cal}}(\mathbf{z}_{\text{priv}}),
\end{equation}

This refined representation $\hat{\mathbf{z}}_{\text{priv}}$ is processed by the tail model $\mathcal{M}_{\text{tail}}$ to compute the model's final layer representation and the autoregressive likelihood over the target text $y^{*} = [y^{*}_1, y^{*}_2, \dots, y^{*}_T] \in \mathcal{Y}_{\text{priv}}$.
This representation is then used to compute the loss function 
\begin{align}
p(\cdot \mid y^{*}_{<t}, \mathbf{z}_{\text{priv}}) &= \mathrm{softmax}\Big(g_3(\mathbf{z}_{\text{priv}}, y^{*}_{<t})\Big), \\
\mathcal{L}_{\mathrm{LLM}} &= -\sum_{t=1}^{T}\log p\!\left(y^{*}_t \mid y^{*}_{<t},\, \mathbf{z}^{*}_{\text{priv}}\right), 
\end{align}

\smallskip
\noindent\textbf{Gradient-perturbed Optimization.}
During this fine-tuning stage, we freeze the parameters of the calibration model $\mathcal{M}_{\text{cal}}$. To further reduce information leakage from the backward pass, we adaptively perturb the gradient calculated by the calibration module before it is sent to the server-side trunk model. Specifically, the client
transmits a perturbed gradient
\begin{equation}
\tilde{\nabla}_{\hat{\mathbf{z}}_{\text{priv}}}\mathcal{L}
=
\nabla_{\hat{\mathbf{z}}_{\text{priv}}}\mathcal{L} + \epsilon,
\quad
\epsilon \sim \mathcal{N}(0, \sigma^2 I),
\end{equation}
where the perturbation scale is determined by the calibration residual:
\begin{equation}
\sigma = \gamma \left\|g_{\text{cal}}(\hat{\mathbf{z}}_{\text{priv}}) -\mathbf{z}^{*}_{\text{priv}} \right\|_2^2 .
\end{equation}
Here, $\gamma$ is a scaling coefficient that controls the strength of the gradient perturbation. Since $\mathcal{M}_{\text{cal}}$ is trained only on the general public or synthesized data, a larger calibration residual indicates a greater deviation from the public calibration distribution and, consequently, a higher likelihood of retaining sample-specific private information. Based on this intuition, \sys applies stronger gradient perturbation to samples with larger residuals. With the perturbed gradient, the server updates the trunk model $\mathcal{M}_{\text{trunk}}$ parameters, and the resulting gradients are further propagated to the client-side head model $\mathcal{M}_{\text{head}}$ following the standard SL backpropagation procedure.

\smallskip
\noindent\textbf{Calibration Model Refinement.}
The calibration model $\mathcal{M}_{\text{cal}}$ is trained to calibrate the representation produced by the initial state of the global model. However, as ($\mathcal{M}_{\text{head}}, \mathcal{M}_{\text{trunk}}, \mathcal{M}_{\text{tail}}$) evolves during fine-tuning, the distribution of the representations from both client and server may drift, which can reduce calibration accuracy. To maintain alignment, we periodically perform a refinement step. Specifically, for every $T_r$ steps, we freeze the global model parameters to update $\mathcal{M}_{\text{cal}}$ using the same public-data supervision procedure described in Section \ref{sec:step-1}. This alternating optimization strategy aims to stabilize the refinement quality throughout private fine-tuning. We present the detailed procedure of privacy-preserving fine-tuning in Algorithm \ref{alg:private_finetuning}.
\begin{algorithm}[t]
\scriptsize
\caption{Privacy-Preserving Fine-tuning}
\label{alg:private_finetuning}
\KwIn{
Private dataset $\mathcal{D}_{\text{priv}}$, public dataset $\mathcal{D}_{\text{pub}}$,
support set $\mathcal{X}_{\text{supp}}$,
head model $g_1$, trunk model $g_2$, tail model $g_3$,
calibration model $g_{\text{cal}}$,
source size $k$, mixture size $m$,
refinement interval $T_{\text{ref}}$, perturbation
coefficient $\gamma$
}
\KwOut{
Fine-tuned model $(g_1,g_2,g_3)$ and calibration model $g_{\text{cal}}$
}

\SetKwFunction{MixupSL}{MixupSL}
\SetKwFunction{Calib}{CalibrationTraining}

Sample secret token pattern $s$;

$\mathcal{X}^{\dagger}_{\text{priv}}\leftarrow
\{\Phi(x;s)\mid x\in \mathcal{X}_{\text{priv}}\}$,
$\mathcal{X}^{\dagger}_{\text{supp}}\leftarrow\{\Phi(x;s)\mid x\in \mathcal{X}_{\text{supp}}\}$

\ForEach{training step $t$ with batch $(x^{*}_{\text{priv}},y^{*}_{\text{priv}})$}{

    Freeze $g_{\text{cal}}$\;

    $\hat{\mathbf{z}}_{\text{priv}} \leftarrow$
    \MixupSL{$x^{*}_{\text{priv}}$, $\mathcal{X}_{\text{supp}}$, $g_1$, $g_2$, $k$, $m$}

    $\mathbf{z}_{\text{priv}}\leftarrow g_{\text{cal}}(\hat{\mathbf{z}}_{\text{priv}})$
    \tcp*[r]{calibrated representation}

    $\hat{y}\leftarrow g_3(\mathbf{z}_{\text{priv}})$
    \tcp*[r]{tail prediction}

    $\mathcal{L}_{\text{LLM}}\leftarrow
    \ell(\hat{y},y^{*}_{\text{priv}})$\;

    Compute the backward gradient $\nabla_{\hat{\mathbf{z}}_{\text{priv}}}\mathcal{L}_{\text{LLM}}$ through $g_{\text{cal}}$\;
    $r_t \leftarrow \|\hat{\mathbf{z}}_{\text{priv}}-\mathbf{z}_{\text{priv}}\|_2^2$\tcp*[r]{calibration residual}
    
    $\sigma_t \leftarrow \gamma r_t$\;

    Sample perturbation $\epsilon_t \sim \mathcal{N}(0,\sigma_t^2 I)$\;

    $\widetilde{\nabla}_{\hat{\mathbf{z}}_{\text{priv}}}\mathcal{L}_{\text{LLM}}\leftarrow \nabla_{\hat{\mathbf{z}}_{\text{priv}}}\mathcal{L}_{\text{LLM}}+\epsilon_t$\;

    Update $g_2$ using $\widetilde{\nabla}_{\hat{\mathbf{z}}_{\text{priv}}}\mathcal{L}_{\text{LLM}}$, and propagate the resulting gradients to update $g_1$\;

    \If{$t \bmod T_r = 0$}{
        Freeze $g_1$, $g_2$, and $g_3$\;

        $g_{\text{cal}}\leftarrow$
        \Calib{$\mathcal{D}_{\text{pub}}$, $\mathcal{X}_{\text{supp}}$, $g_1$, $g_2$, $g_{\text{cal}}$, $k$, $m$}
        \tcp*[r]{calibration refinement}

        Unfreeze $g_1$, $g_2$, and $g_3$; Freeze $g_{\text{cal}}$ \; 
    }
}

\Return{$(g_1,g_2,g_3), g_{\text{cal}}$}\;
\end{algorithm}

\subsection{Inference and Deployment}
After training, \sys can be deployed in a fully local or a split inference mode. For fully local inference, the client obtains all model segments and performs end-to-end prediction locally, which removes communication overhead during inference. For split inference, the client retains the head and tail models, while the server continues to host the trunk model as in the training stage. This setting preserves the computational advantage of SL for resource-limited clients.

In both settings, the calibration model can be optionally inserted between the trunk and the tail model. Enabling calibration refines the intermediate representation, whereas disabling it reduces inference-time overhead. As a result, \sys supports flexible deployment choices that can be selected according to practical requirements on utility, latency, and communication. In the experiments, we evaluate both inference variants under settings with and without calibration.

\begin{table*}[t]
\centering
\scriptsize
\setlength{\tabcolsep}{3.8pt}
\renewcommand{\arraystretch}{0.8}
\caption{Performance of Utility and against Data Reconstruction Attack on classification tasks.}
\begin{tabular}{ll|ccccccccc|ccccccccc}
\toprule
\multirow{3}{*}{Dataset} & \multirow{3}{*}{Baseline}
& \multicolumn{9}{c|}{Qwen3-1.7B}
& \multicolumn{9}{c}{Llama 3.2-1B} \\
\cmidrule(lr){3-11}
\cmidrule(lr){12-20}
& 
& \multicolumn{1}{c}{Utility}
& \multicolumn{2}{c}{SIP}
& \multicolumn{2}{c}{TAG}
& \multicolumn{2}{c}{LAMP}
& \multicolumn{2}{c|}{BiSR}
& \multicolumn{1}{c}{Utility}
& \multicolumn{2}{c}{SIP}
& \multicolumn{2}{c}{TAG}
& \multicolumn{2}{c}{LAMP}
& \multicolumn{2}{c}{BiSR} \\
\cmidrule(lr){3-3}
\cmidrule(lr){4-5}
\cmidrule(lr){6-7}
\cmidrule(lr){8-9}
\cmidrule(lr){10-11}
\cmidrule(lr){12-12}
\cmidrule(lr){13-14}
\cmidrule(lr){15-16}
\cmidrule(lr){17-18}
\cmidrule(lr){19-20}
& 
& ACC$\uparrow$ & RF$\downarrow$ & ME$\downarrow$ & RF$\downarrow$ & ME$\downarrow$ & RF$\downarrow$ & ME$\downarrow$ & RF$\downarrow$ & ME$\downarrow$
& ACC$\uparrow$ & RF$\downarrow$ & ME$\downarrow$ & RF$\downarrow$ & ME$\downarrow$ & RF$\downarrow$ & ME$\downarrow$ & RF$\downarrow$ & ME$\downarrow$ \\
\midrule

\multirow{6}{*}{CoLA}
& Non-split & 0.838 & 0.622 & 0.585 & 0.452 & 0.426 & 0.484 & 0.436 & 0.116 & 0.155 & 0.845 & 0.548 & 0.534 & 0.430 & 0.328 & 0.433 & 0.327 & 0.193 & 0.183 \\
& DP-SGD & 0.845 & 0.573 & 0.575 & 0.476 & 0.429 & 0.480 & 0.433 & 0.118 & 0.154 & 0.850 & 0.530 & 0.503 & 0.418 & 0.321 & 0.418 & 0.323 & 0.196 & 0.173 \\
& DP-forward & 0.786 & 0.392 & 0.359 & 0.387 & 0.380 & 0.375 & 0.365 & 0.150 & 0.137 & 0.838 & 0.530 & 0.504 & 0.461 & 0.410 & 0.465 & 0.415 & 0.282 & 0.204 \\
& SnD & 0.309 & 0.604 & 0.563 & 0.269 & 0.233 & 0.266 & 0.229 & 0.112 & 0.081 & 0.691 & 0.239 & 0.167 & 0.481 & 0.463 & 0.483 & 0.463 & 0.242 & 0.243 \\
& Naive Mix & 0.498 & 0.456 & 0.415 & 0.390 & 0.345 & 0.388 & 0.344 & \textbf{0.076} & \textbf{0.073} & 0.742 & 0.339 & 0.382 & 0.409 & 0.332 & 0.414 & 0.342 & 0.212 & 0.194 \\
& Ours & \textbf{0.846} & \textbf{0.135} &\textbf{ 0.128} &\textbf{ 0.240} & \textbf{0.193} & \textbf{0.251} & \textbf{0.196} & 0.092 & 0.081 & \textbf{0.839} & \textbf{0.019} & \textbf{0.016} & \textbf{0.183} & \textbf{0.113} & \textbf{0.170} & \textbf{0.109} & \textbf{0.127} & \textbf{0.078} \\
\midrule

\multirow{6}{*}{SST-2}
& Non-split & 0.956 & 0.672 & 0.690 & 0.417 & 0.308 & 0.416 & 0.298 & 0.176 & 0.169 & 0.963 & 0.657 & 0.659 & 0.409 & 0.268 & 0.411 & 0.265 & 0.178 & 0.151 \\
& DP-SGD & 0.958 & 0.680 & 0.707 & 0.411 & 0.304 & 0.412 & 0.303 & 0.177 & 0.149 & 0.960 & 0.580 & 0.580 & 0.403 & 0.258 & 0.402 & 0.258 & 0.198 & 0.206 \\
& DP-forward & \textbf{0.971} & 0.485 & 0.481 & 0.391 & 0.280 & 0.382 & 0.258 & 0.173 & 0.141 & 0.953 & 0.491 & 0.489 & 0.378 & 0.271 & 0.378 & 0.272 & 0.215 & 0.215 \\
& SnD & 0.491 & 0.679 & 0.685 & 0.203 & 0.161 & 0.200 & 0.158 & 0.116 & 0.092 & 0.509 & 0.654 & 0.627 & 0.426 & 0.374 & 0.425 & 0.370 & 0.181 & 0.180 \\
& Naive Mix & 0.796 & 0.444 & 0.484 & 0.413 & 0.343 & 0.445 & 0.357 & 0.154 & 0.145 & \textbf{0.964} & 0.473 & 0.443 & 0.387 & 0.276 & 0.409 & 0.280 & 0.169 & 0.138 \\
& Ours & 0.952 & \textbf{0.127} & \textbf{0.102} &\textbf{ 0.195} & \textbf{0.118} & \textbf{0.185} & \textbf{0.111} & \textbf{0.098} & \textbf{0.060} & 0.953 & \textbf{0.107} & \textbf{0.071} & \textbf{0.145} & \textbf{0.071} & \textbf{0.133} & \textbf{0.068} & \textbf{0.139} & \textbf{0.100} \\
\midrule

\multirow{6}{*}{MRPC}
& Non-split & 0.863 & 0.772 & 0.789 & 0.583 & 0.496 & 0.595 & 0.493 & 0.277 & 0.243 & 0.860 & 0.759 & 0.778 & 0.600 & 0.509 & 0.603 & 0.511 & 0.342 & 0.276 \\
& DP-SGD & 0.831 & 0.761 & 0.781 & 0.587 & 0.492 & 0.598 & 0.493 & 0.279 & 0.238 & \textbf{0.860} & 0.787 & 0.798 & 0.605 & 0.522 & 0.603 & 0.519 & 0.333 & 0.252 \\
& DP-forward & 0.794 & 0.595 & 0.612 & 0.528 & 0.436 & 0.543 & 0.419 & 0.304 & 0.232 & 0.853 & 0.627 & 0.646 & 0.519 & 0.383 & 0.516 & 0.374 & 0.308 & 0.235 \\
& SnD & 0.684 & 0.783 & 0.797 & 0.453 & 0.424 & 0.467 & 0.430 & 0.299 & 0.250 & 0.684 & 0.732 & 0.742 & 0.613 & 0.545 & 0.612 & 0.542 & 0.328 & 0.274 \\
& Naive Mix & 0.748 & 0.507 & 0.487 & 0.549 & 0.470 & 0.481 & 0.399 & 0.255 & 0.203 & 0.831 & 0.345 & 0.326 & 0.358 & 0.289 & 0.504 & 0.416 & 0.202 & 0.156 \\
& Ours & \textbf{0.880} & \textbf{0.069} & \textbf{0.038} & \textbf{0.224} & \textbf{0.135} & \textbf{0.232} &\textbf{0.147} & \textbf{0.098} & \textbf{0.051} & 0.836 & \textbf{0.069} & \textbf{0.036} & \textbf{0.147} & \textbf{0.076} & \textbf{0.113} & \textbf{0.052} & \textbf{0.102} & \textbf{0.058} \\
\midrule

\multirow{6}{*}{RTE}
& Non-split & 0.870 & 0.758 & 0.748 & 0.509 & 0.455 & 0.504 & 0.443 & 0.226 & 0.205 & 0.783 & 0.709 & 0.707 & 0.491 & 0.430 & 0.496 & 0.433 & 0.283 & 0.244 \\
& DP-SGD & 0.610 & 0.750 & 0.753 & 0.505 & 0.447 & 0.502 & 0.439 & 0.233 & 0.210 & 0.779 & 0.715 & 0.711 & 0.495 & 0.435 & 0.523 & 0.434 & 0.279 & 0.239 \\
& DP-forward & 0.585 & 0.542 & 0.544 & 0.380 & 0.323 & 0.357 & 0.296 & 0.215 & 0.188 & 0.740 & 0.601 & 0.604 & 0.445 & 0.339 & 0.446 & 0.349 & 0.291 & 0.254 \\
& SnD & 0.529 & 0.755 & 0.762 & 0.389 & 0.362 & 0.385 & 0.360 & 0.278 & 0.242 & 0.471 & 0.694 & 0.719 & 0.496 & 0.446 & 0.496 & 0.446 & 0.249 & 0.210 \\
& Naive Mix & 0.715 & 0.457 & 0.464 & 0.527 & 0.474 & 0.392 & 0.358 & 0.206 & 0.178 & 0.726 & 0.428 & 0.428 & 0.457 & 0.401 & 0.390 & 0.344 & 0.229 & 0.189 \\
& Ours & \textbf{0.866} & \textbf{0.059} & \textbf{0.035} & \textbf{0.185} & \textbf{0.121} & \textbf{0.190} & \textbf{0.127} & \textbf{0.105} & \textbf{0.064} & \textbf{0.783} & \textbf{0.061} & \textbf{0.044} & \textbf{0.150} & \textbf{0.089} & \textbf{0.112} & \textbf{0.064} & \textbf{0.111} & \textbf{0.059} \\
\bottomrule
\end{tabular}
\label{tab:performance-classification}
\end{table*}

\section{Evaluation}
\subsection{Experimental Setup}
\textbf{Dataset}. We evaluate our method on both classification and generation tasks. For classification, we use four representative datasets from the GLUE benchmark \cite{wang2018glue}: CoLA, SST-2, QQP, and RTE: CoLA assesses linguistic acceptability (grammaticality judgments) \cite{warstadt2019neural}, SST-2 is a binary sentiment classification task \cite{socher2013recursive}, QQP measures semantic equivalence between question pairs, and RTE evaluates textual entailment in a low-resource natural language inference setting. 
We further evaluate on four generative benchmarks. GSM8K is used to evaluate multi-step math word problem solving \cite{cobbe2021training}. CodeAlpaca-20K is adopted for code generation under instruction-following prompts \cite{chaudhary2023code}. SAMSum is a dialogue summarization benchmark, evaluating the model's ability to generate concise summaries from multi-turn conversational contexts \cite{gliwa2019samsum}. WebNLG is included as a data-to-text generation benchmark, where structured RDF-style input triples are converted into natural language descriptions \cite{gardent2017webnlg}. In addition, we construct the public and support datasets from Wikitext \cite{merity2016pointer}, Yelp \cite{zhang2015character}, and Multi-Source \cite{mai2023split}.

\textbf{Target LLMs.} Our main experiments are conducted on Qwen3-1.7B \cite{yang2025qwen3} and Llama 3.2-1B \cite{grattafiori2024llama}, which represent two widely adopted open-sourced LLM families. To study the effect of model scale, we further evaluate \sys on Qwen3-4B, Qwen3-8B, and Qwen3-14B. We also extend our evaluation to different model architectures, including Qwen3-30B-A3B, a Mixture-of-Experts (MoE) LLM. In addition to full-parameter fine-tuning, we assess the effectiveness of \sys under parameter-efficient fine-tuning (PEFT) settings using Low-Rank Adaptation (LoRA) \cite{hu2022lora}. 


\textbf{Baselines.} We compare \sys against representative privacy defenses for SL. In addition to the standard non-split baseline, we include DP-Forward \cite{du2023dp}, DP-SGD \cite{abadi2016deep}, and Split-and-Denoise (SnD) \cite{mai2023split}, which protect private information by introducing noise into the intermediate exchanged signals. We additionally include Naive Mix \cite{sun2020mixup}, a naive representation-mixing baseline that is more closely related to our setting.

\textbf{Attacks.} \sys is evaluated against four DRAs in the SL setting, i.e., SIP, TAG, LAMP, and BiSR. SIP is a forward smashed-data inversion attack, where the adversary trains an inversion model on auxiliary data and then uses the observed smashed representations to reconstruct the original input text \cite{chen2024unveiling}. TAG and LAMP are backward gradient inversion attacks that exploit the gradients propagated through the split model to recover private supervision signals \cite{deng2021tag, balunovic2022lamp}. Finally, BiSR is a stronger bidirectional reconstruction attack that combines forward inversion and backward gradient matching, and reconstructs private data from both communication directions \cite{chen2024unveiling}. 

\textbf{Evaluation Metrics.} We report the default evaluation metrics for each dataset. For the GLUE classification tasks, including CoLA, SST-2, QQP, and RTE, we report Accuracy (ACC). For the generative tasks, we evaluate GSM8K using answer accuracy under exact matching (EMA) \cite{cobbe2021training}, SAMSum using ROUGE-L F1 (RF) \cite{lin2004rouge}, WebNLG using BLEU \cite{papineni2002bleu}, and CodeAlpaca-20K using CodeBLEU \cite{ren2020codebleu}. To assess the effectiveness of DRAs, we report ROUGE-L F1 (RF) \cite{lin2004rouge} and METEOR (ME) \cite{banerjee2005meteor}, which measure the longest common subsequence (LCS) and lexical-semantic similarity between the reference and reconstructed text, respectively. 

We provide additional details on the evaluation metrics and implementation settings in Appendix \ref{sec:experimental-details}.

\subsection{Main Results}
\textbf{Classification Tasks.} We first evaluate \sys on four GLUE classification benchmarks. Table \ref{tab:performance-classification} reports the downstream utility and reconstruction results. 

Overall, \sys preserves downstream utility across both models. On Llama 3.2-1B, \sys achieves an accuracy of 0.839 on CoLA and 0.953 on SST-2, compared with 0.845 and 0.963 under non-split training, respectively. The corresponding utility gaps are limited to less than 1\%. A similar trend is observed on Qwen3-1.7B. \sys reaches 0.846 accuracy on CoLA, slightly exceeding the non-split baseline of 0.838, while maintaining comparable performance on SST-2 (0.952 vs. 0.956). On MRPC and RTE, \sys achieves accuracies of 0.880 and 0.866 on Qwen3-1.7B, respectively, which are also close to the corresponding non-split baselines of 0.863 and 0.870. These results indicate that the proposed obfuscation procedure effectively retains task-relevant signals for the classification and avoids model utility degradation.

\begin{table*}[t]
\centering
\scriptsize
\setlength{\tabcolsep}{3pt}
\renewcommand{\arraystretch}{0.8}
\caption{Performance of Utility and against Data Reconstruction Attack on generation tasks.}
\begin{tabular}{cl|ccccccccc|ccccccccc}
\toprule
\multirow{2}{*}{Dataset} & \multirow{2}{*}{Baseline}
& \multicolumn{9}{c|}{Qwen3-1.7B}
& \multicolumn{9}{c}{Llama 3.2-1B} \\
\cmidrule(lr){3-11}
\cmidrule(lr){12-20}
& 
& \multicolumn{1}{c}{Utility}
& \multicolumn{2}{c}{SIP}
& \multicolumn{2}{c}{TAG}
& \multicolumn{2}{c}{LAMP}
& \multicolumn{2}{c|}{BiSR}
& \multicolumn{1}{c}{Utility}
& \multicolumn{2}{c}{SIP}
& \multicolumn{2}{c}{TAG}
& \multicolumn{2}{c}{LAMP}
& \multicolumn{2}{c}{BiSR} \\
\midrule
& 
& EMA$\uparrow$ & RF$\downarrow$ & ME$\downarrow$ & RF$\downarrow$ & ME$\downarrow$ & RF$\downarrow$ & ME$\downarrow$ & RF$\downarrow$ & ME$\downarrow$
& EMA$\uparrow$ & RF$\downarrow$ & ME$\downarrow$ & RF$\downarrow$ & ME$\downarrow$ & RF$\downarrow$ & ME$\downarrow$ & RF$\downarrow$ & ME$\downarrow$ \\
\cmidrule(lr){3-3}
\cmidrule(lr){4-5}
\cmidrule(lr){6-7}
\cmidrule(lr){8-9}
\cmidrule(lr){10-11}
\cmidrule(lr){12-12}
\cmidrule(lr){13-14}
\cmidrule(lr){15-16}
\cmidrule(lr){17-18}
\cmidrule(lr){19-20}

\multirow{3}{*}{\makecell{GSM8K}}
& Non-split & 0.469 & 0.760 & 0.782 & 0.743 & 0.840 & 0.744 & 0.839 & 0.424 & 0.347 & 0.254 & 0.698 & 0.700 & 0.746 & 0.821 & 0.757 & 0.825 & 0.344 & 0.294 \\
& DP-SGD & 0.461 & 0.738 & 0.776 & 0.742 & 0.839 & 0.743 & 0.837 & 0.418 & 0.344 & 0.187 & 0.585 & 0.567 & 0.485 & 0.593 & 0.450 & 0.583 & 0.212 & 0.210 \\
& DP-forward & \textbf{0.467} & 0.570 & 0.565 & 0.740 & 0.830 & 0.736 & 0.835 & 0.398 & 0.359 & 0.232 & 0.585 & 0.566 & 0.748 & 0.822 & 0.756 & 0.826 & 0.357 & 0.290 \\
& Naive Mix & 0.461 & 0.461 & 0.426 & 0.481 & 0.409 & 0.551 & 0.483 & 0.299 & 0.244 & 0.225 & 0.387 & 0.358 & 0.578 & 0.564 & 0.521 & 0.450 & 0.277 & 0.211 \\
& Ours & 0.455 & \textbf{0.042} & \textbf{0.024} & \textbf{0.303} & \textbf{0.349} & \textbf{0.272} & \textbf{0.305} & \textbf{0.106} & \textbf{0.066} & \textbf{0.236} & \textbf{0.037} & \textbf{0.026} & \textbf{0.303} & \textbf{0.346} & \textbf{0.273} & \textbf{0.305} & \textbf{0.174} & \textbf{0.140} \\
\midrule
& 
& BLEU$\uparrow$ & RF$\downarrow$ & ME$\downarrow$ & RF$\downarrow$ & ME$\downarrow$ & RF$\downarrow$ & ME$\downarrow$ & RF$\downarrow$ & ME$\downarrow$
& BLEU$\uparrow$ & RF$\downarrow$ & ME$\downarrow$ & RF$\downarrow$ & ME$\downarrow$ & RF$\downarrow$ & ME$\downarrow$ & RF$\downarrow$ & ME$\downarrow$ \\
\cmidrule(lr){3-3}
\cmidrule(lr){4-5}
\cmidrule(lr){6-7}
\cmidrule(lr){8-9}
\cmidrule(lr){10-11}
\cmidrule(lr){12-12}
\cmidrule(lr){13-14}
\cmidrule(lr){15-16}
\cmidrule(lr){17-18}
\cmidrule(lr){19-20}

\multirow{3}{*}{\makecell{WebNLG}}
& Non-split & 43.22 & 0.663 & 0.630 & 0.700 & 0.644 & 0.703 & 0.644 & 0.304 & 0.249 & 42.84 & 0.568 & 0.542 & 0.715 & 0.621 & 0.716 & 0.620 & 0.313 & 0.294 \\
& DP-SGD & 41.17 & 0.658 & 0.618 & 0.378 & 0.358 & 0.378 & 0.359 & 0.337 & 0.287 & 0.000 & 0.574 & 0.527 & 0.414 & 0.354 & 0.410 & 0.355 & \textbf{0.105} & \textbf{0.080} \\
& DP-forward & \textbf{43.11} & 0.439 & 0.424 & 0.687 & 0.612 & 0.691 & 0.627 & 0.310 & 0.261 & 41.49 & 0.464 & 0.424 & 0.682 & 0.599 & 0.696 & 0.599 & 0.315 & 0.268 \\

& Naive Mix & 42.25 & 0.368 & 0.324 & 0.478 & 0.308 & 0.432 & 0.293 & 0.223 & 0.155 & 40.76 & 0.300 & 0.254 & 0.483 & 0.379 & 0.402 & 0.301 & 0.231 & 0.188 \\
& Ours & 39.65 & \textbf{0.082} & \textbf{0.030} & \textbf{0.292} & \textbf{0.278} & \textbf{0.292} & \textbf{0.261} & \textbf{0.112} & \textbf{0.067} & \textbf{41.66} & \textbf{0.052} & \textbf{0.019} & \textbf{0.186} & \textbf{0.169} & \textbf{0.148} & \textbf{0.132} & 0.116 & 0.091 \\
\midrule
& 
& RF$\uparrow$ & RF$\downarrow$ & ME$\downarrow$ & RF$\downarrow$ & ME$\downarrow$ & RF$\downarrow$ & ME$\downarrow$ & RF$\downarrow$ & ME$\downarrow$
& RF$\uparrow$ & RF$\downarrow$ & ME$\downarrow$ & RF$\downarrow$ & ME$\downarrow$ & RF$\downarrow$ & ME$\downarrow$ & RF$\downarrow$ & ME$\downarrow$ \\
\cmidrule(lr){3-3}
\cmidrule(lr){4-5}
\cmidrule(lr){6-7}
\cmidrule(lr){8-9}
\cmidrule(lr){10-11}
\cmidrule(lr){12-12}
\cmidrule(lr){13-14}
\cmidrule(lr){15-16}
\cmidrule(lr){17-18}
\cmidrule(lr){19-20}

\multirow{3}{*}{\makecell{SAMSum}}
& Non-split & 0.534 & 0.745 & 0.678 & 0.549 & 0.516 & 0.550 & 0.517 & 0.272 & 0.217 & 0.524 & 0.696 & 0.630 & 0.518 & 0.488 & 0.519 & 0.486 & 0.255 & 0.214 \\

& DP-SGD & 0.528 & 0.751 & 0.680 & 0.296 & 0.206 & 0.339 & 0.201 & 0.315 & 0.301 & 0.000 & 0.706 & 0.647 & 0.209 & 0.197 & 0.207 & 0.194 & \textbf{0.111} & \textbf{0.112} \\
& DP-forward & 0.536 & 0.758 & 0.678 & 0.523 & 0.519 & 0.528 & 0.525 & 0.313 & 0.240 & \textbf{0.524} & 0.565 & 0.494 & 0.520 & 0.486 & 0.518 & 0.486 & 0.294 & 0.230 \\
& Naive Mix & \textbf{0.539} & 0.392 & 0.335 & 0.421 & 0.407 & 0.500 & 0.477 & 0.212 & 0.166 & 0.519 & 0.089 & 0.049 & 0.384 & 0.266 & 0.401 & 0.280 & 0.315 & 0.222 \\
& Ours & 0.532 & \textbf{0.156} & \textbf{0.098} & \textbf{0.135} & \textbf{0.141} & \textbf{0.142} & \textbf{0.135} & \textbf{0.099} & \textbf{0.064} & 0.517 & \textbf{0.078} & \textbf{0.044} & \textbf{0.153} & \textbf{0.168} & \textbf{0.149} & \textbf{0.157} & 0.151 & 0.133 \\
\midrule
& 
& {\tiny CodeBLEU}$\uparrow$ & RF$\downarrow$ & ME$\downarrow$ & RF$\downarrow$ & ME$\downarrow$ & RF$\downarrow$ & ME$\downarrow$ & RF$\downarrow$ & ME$\downarrow$
& {\tiny CodeBLEU}$\uparrow$ & RF$\downarrow$ & ME$\downarrow$ & RF$\downarrow$ & ME$\downarrow$ & RF$\downarrow$ & ME$\downarrow$ & RF$\downarrow$ & ME$\downarrow$ \\
\cmidrule(lr){3-3}
\cmidrule(lr){4-5}
\cmidrule(lr){6-7}
\cmidrule(lr){8-9}
\cmidrule(lr){10-11}
\cmidrule(lr){12-12}
\cmidrule(lr){13-14}
\cmidrule(lr){15-16}
\cmidrule(lr){17-18}
\cmidrule(lr){19-20}
\multirow{3}{*}{\makecell{CodeAlpaca}}
& Non-split & 0.536 & 0.630 & 0.615 & 0.779 & 0.809 & 0.777 & 0.811 & 0.379 & 0.308 & 0.514 & 0.545 & 0.536 & 0.783 & 0.811 & 0.783 & 0.810 & 0.384 & 0.344 \\
& DP-SGD & 0.536 & 0.633 & 0.610 & 0.761 & 0.806 & 0.770 & 0.808 & 0.356 & 0.307 & 0.401 & 0.539 & 0.544 & 0.481 & 0.546 & 0.479 & 0.546 & 0.341 & 0.315 \\

& DP-forward & 0.530 & 0.427 & 0.409 & 0.801 & 0.807 & 0.790 & 0.797 & 0.374 & 0.294 & 0.514 & 0.487 & 0.486 & 0.777 & 0.799 & 0.779 & 0.805 & 0.389 & 0.335 \\
& Naive Mix & 0.534 & 0.372 & 0.338 & 0.503 & 0.464 & 0.393 & 0.348 & 0.251 & 0.192 & \textbf{0.518} & 0.296 & 0.252 & 0.524 & 0.463 & 0.524 & 0.483 & 0.247 & 0.183 \\
& Ours & \textbf{0.543} & \textbf{0.059} & \textbf{0.031} & \textbf{0.309} & \textbf{0.296} & \textbf{0.271} & \textbf{0.245} & \textbf{0.143} & \textbf{0.077} & 0.502 & \textbf{0.073} & \textbf{0.030} & \textbf{0.308} & \textbf{0.281} & \textbf{0.303} & \textbf{0.308} & \textbf{0.245} & \textbf{0.184} \\
\bottomrule
\end{tabular}
\label{tab:performance-generation}
\end{table*}

\sys also substantially suppresses reconstruction across DRAs that exploit forward representations, backward gradients, or both. For example, on CoLA with Llama 3.2-1B, \sys reduces the SIP reconstruction scores from 0.548/0.534 under non-split training to 0.019/0.016 in terms of RF/ME. Under TAG and LAMP, \sys limits the reconstruction scores to 0.183/0.113 and 0.170/0.109, respectively, compared with 0.461/0.410 and 0.465/0.415 under DP-Forward. \sys is also effective against BiSR, which combines forward inversion and gradient matching. The BiSR scores on CoLA are 0.127/0.078, substantially lower than the 0.282/0.204 obtained by DP-Forward and 0.242/0.243 by SnD. The same pattern is observed on Qwen3-1.7B. For instance, on RTE, \sys reduces the SIP score from 0.758/0.748 under non-split training to 0.059/0.035 and limits the TAG score to 0.185/0.121, whereas the other defenses provide little or no reduction.

\textbf{Generation Tasks.} We further evaluate \sys on four generation tasks. The results are reported in Table \ref{tab:performance-generation}.

\sys maintains competitive generation quality across most evaluated settings. On Llama 3.2-1B, \sys achieves 0.236 EMA on GSM8K, 41.66 BLEU on WebNLG, 0.517 RF on SAMSum, and 0.502 CodeBLEU on CodeAlpaca. These results are close to the corresponding non-split training results of 0.254, 42.84, 0.524, and 0.514, respectively. The performance is extended to Qwen3-1.7B. For example, \sys achieves 0.455 EMA on GSM8K, 0.532 RF on SAMSum, and 0.543 CodeBLEU on CodeAlpaca, closely matching or slightly exceeding the corresponding non-split results of 0.469, 0.534, and 0.536. WebNLG represents a more challenging setting for Qwen3-1.7B, where the BLEU score decreases from 43.22 to 39.65. This performance gap suggests that structured data-to-text generation may be more sensitive to secret-token obfuscation or representation approximation errors.

Generation tasks typically involve longer output sequences, thereby exposing a broader attack surface to DRAs. Despite this challenge, \sys consistently suppresses the effectiveness of DRAs. On GSM8K with Qwen3-1.7B, \sys reduces the SIP reconstruction scores from 0.760/0.782 under non-split training to 0.042/0.024, corresponding to around 95\% relative decrease. It also lowers TAG from 0.743/0.840 to 0.303/0.349 and BiSR from 0.424/0.347 to 0.106/0.066. Similarly, on CodeAlpaca, \sys limits SIP to 0.059/0.031 and BiSR to 0.143/0.077, compared with 0.630/0.615 and 0.379/0.308 under non-split training. These results demonstrate that the proposed mixup-based obfuscation mechanisms are also effective against DRAs on more challenging generation tasks.

\smallskip
\noindent\textbf{Cross-task Stability.} Notably, the baseline comparison further highlights the stability of \sys across diverse tasks. Existing defenses often incur unstable downstream utility and, in some settings, lead to model collapse. For example, on CoLA with Qwen3-1.7B, SnD and NaiveMix achieve accuracies of only 0.309 and 0.498, respectively, whereas \sys reaches 0.846. Similarly, on WebNLG with Llama 3.2-1B, DP-SGD collapses to a BLEU score of 0, while \sys maintains this value at 41.66. Across datasets, LLM families, and DRA variants, \sys consistently preserves stable training outcome while achieving a strong privacy–utility trade-off. We provide a more detailed reproducibility analysis of \sys in Appendix \ref{sec:reproducibility}.

\smallskip
\noindent\textbf{Generalization Across Model Scales, Architectures, and Fine-Tuning Strategies.}
\begin{figure}[t]
    \centering
    \includegraphics[width=\columnwidth]{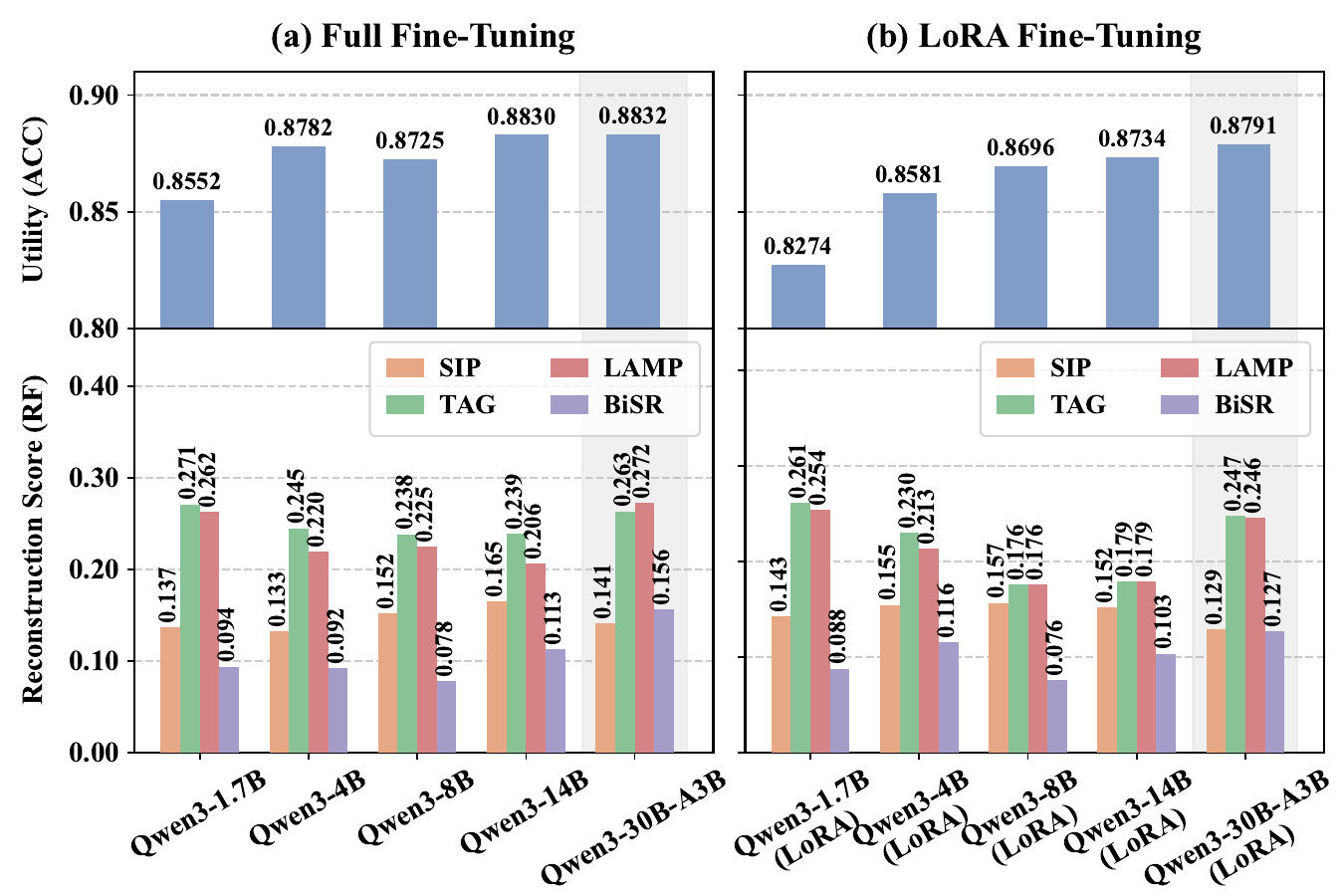}
    \caption{Performance of \sys across model scales, architectures, and fine-tuning strategies.} 
    \label{fig:more-models}
\end{figure}

We further evaluate \sys across diverse LLM and fine-tuning configurations. Figure \ref{fig:more-models} reports the results on CoLA using Qwen3 models with 1.7B, 4B, 8B, and 14B parameters, and the Qwen3-30B-A3B MoE model.

As shown in Figure \ref{fig:more-models}(a), full fine-tuning with \sys on larger models generally improves the downstream utility, which is consistent with the trend in the non-split setting. However, the reconstruction results exhibit different patterns across DRAs. Specifically, SIP maintains low reconstruction scores, but shows a modest upward trend as the model size increases. This is likely because larger models produce richer hidden representations, which may preserve slightly more recoverable semantic information. In contrast, gradient-based DRAs become less effective as the model size increases. For example, the RF score decreases from 0.271 to 0.239 for TAG and from 0.262 to 0.206 for LAMP. This may be because larger models distribute sample-specific information across a higher-dimensional parameter space, which makes the optimization-based gradient inversion more challenging. BiSR remains broadly stable across model sizes, with RF score ranging only from 0.078 to 0.113.

Figure \ref{fig:more-models}(b) demonstrates the results under parameter-efficient fine-tuning (PEFT) with LoRA. Compared with full fine-tuning,  LoRA fine-tuning generally yields lower downstream utility for \sys, particularly on smaller models. However, the performance gap generally narrows as the model size increases. This finding is consistent with the performance gap commonly observed between LoRA and full fine-tuning in standard non-split settings. Importantly, the privacy trends under LoRA closely follow those observed under full fine-tuning. Specifically, SIP and BiSR remain relatively stable, whereas TAG and LAMP become less effective as the model size increases.

Finally, the shaded results in Figure \ref{fig:more-models}(a) and \ref{fig:more-models}(b) demonstrate that the MoE model Qwen3-30B-A3B performs comparably to the evaluated dense models. It achieves the highest utility under both full fine-tuning and LoRA, while its reconstruction scores fall between those of Qwen3-1.7B and Qwen3-4B. These findings suggest that \sys can be effectively applied to MoE-based LLMs. Overall, \sys generalizes well across model scales, architecture, and fine-tuning strategies.

\subsection{Ablation Study}\label{sec:ablation-study}
\begin{figure}[t]
    \centering
    \includegraphics[width=\columnwidth]{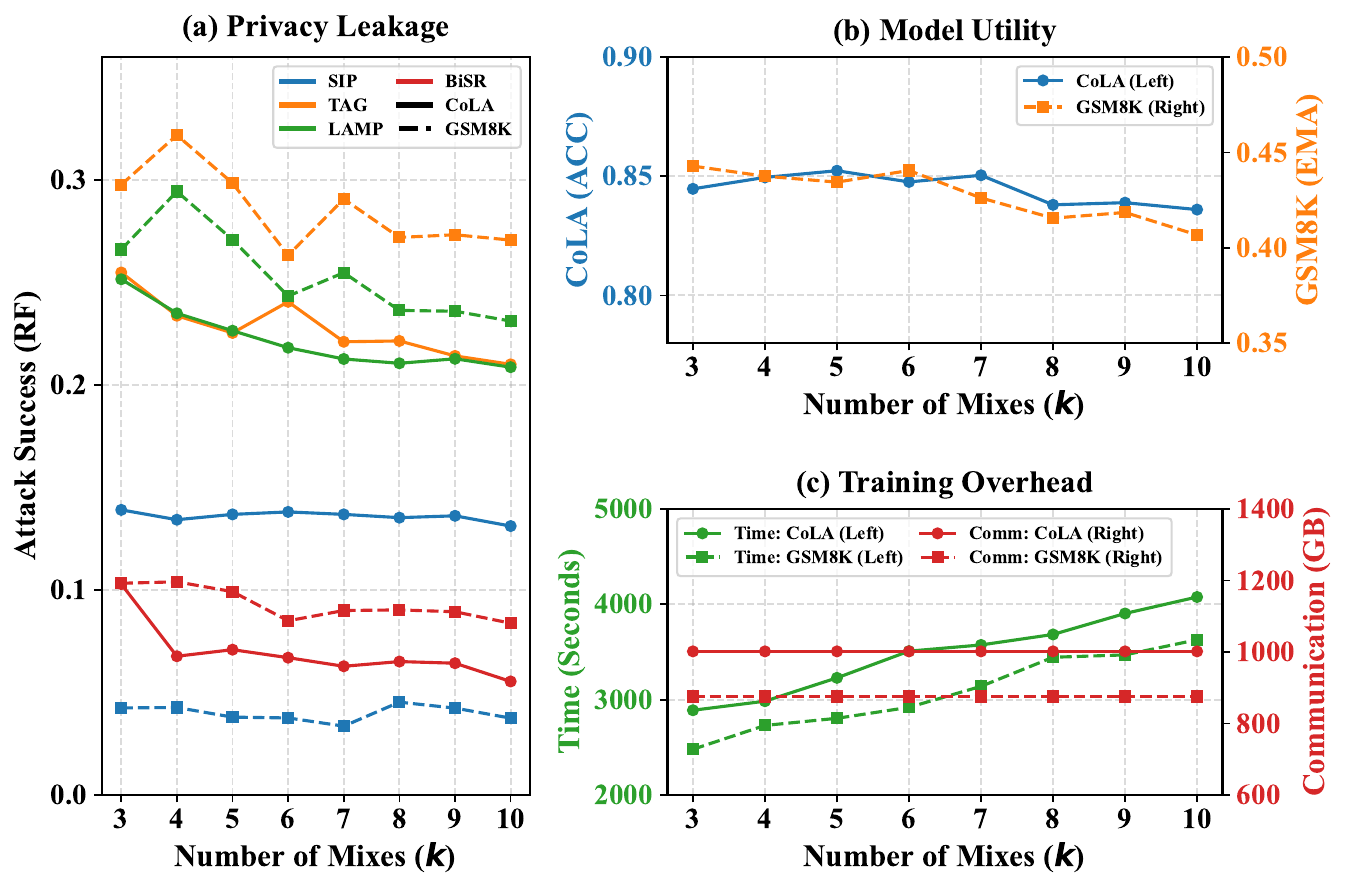}
    \caption{Impact of Source Mixing Size $k$ on CoLA and GSM8K.} 
    \label{fig:varying-k}
\end{figure}

\begin{figure}[t]
    \centering
    \includegraphics[width=\columnwidth]{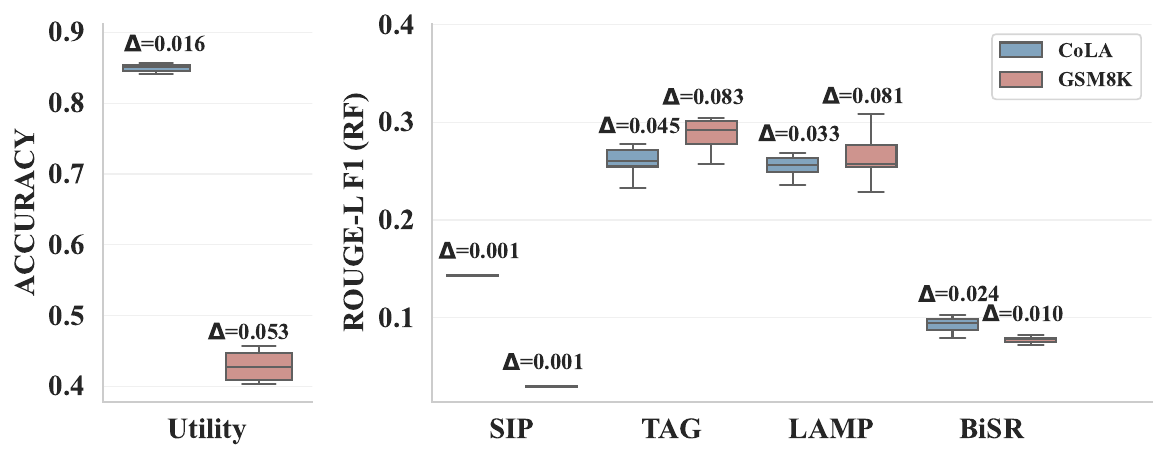}
    \caption{Impact of public and support dataset selection. $\Delta$ denotes the performance range, i.e., $\max - \min$.} 
    \label{fig:dataset}
\end{figure}

We conduct an ablation study to examine the contribution of each module in \sys, with the results reported in Table \ref{tab:ablation-study}, Table \ref{tab:all-ablation-study}, Figure \ref{fig:varying-k}, and Figure \ref{fig:dataset}.

\begin{table}[t]
\centering
\caption{Ablation study on CoLA and GSM8K. N/A: STs are not applied to CoLA.}
\scriptsize
\setlength{\tabcolsep}{1pt}
\renewcommand{\arraystretch}{1}
\begin{tabular}{c l c c c c c c c c c c}
\toprule
\multirow{2}{*}{No.} & \multirow{2}{*}{Setting}
& \multicolumn{5}{c}{CoLA}
& \multicolumn{5}{c}{GSM8K} \\
\cmidrule(lr){3-7} \cmidrule(lr){8-12}
& & Utility & SIP & TAG & LAMP & BiSR
& Utility & SIP & TAG & LAMP & BiSR \\
\midrule
1 & \sys & 0.846 & 0.135 & 0.240 & 0.251 & 0.092 & 0.455 & 0.042 & 0.303 & 0.272 & 0.106 \\
\midrule
\multicolumn{6}{l}{\textbf{Secret Tokens (STs)}} \\
2 & w/o STs
& \multicolumn{5}{c}{\textit{N/A}}
& 0.464 & 0.267 & 0.388 & 0.381 & 0.211 \\
\midrule
\multicolumn{6}{l}{\textbf{Calibration Model (Cal. Model)}} \\
3 &  w/o Cal. Model
& 0.833 & 0.594 & 0.488 & 0.475 & 0.181
& 0.350 & 0.098 & 0.565 & 0.566 & 0.358 \\
4 & w/o Pre-train
& 0.829 & 0.137 & 0.269 & 0.272 & 0.097
& 0.291 & 0.021 & 0.380 & 0.367 & 0.121 \\
\midrule
\multicolumn{6}{l}{\textbf{Adaptive Gradient Perturbation (GP)}} \\
5 & w/o GP
& 0.853 & 0.140 & 0.268 & 0.257 & 0.109
& 0.309 & 0.043 & 0.390 & 0.401 & 0.175 \\

6 & w/ non-adaptive
& 0.852 & 0.141 & 0.280 & 0.278 & 0.103
& 0.306 & 0.042 & 0.370 & 0.376 & 0.157 \\

7 & w/o STs and GP  & \multicolumn{5}{c}{\textit{N/A}} & 0.479 & 0.266 & 0.475 & 0.468 & 0.255 \\
\bottomrule
\end{tabular}
\label{tab:ablation-study}
\end{table}

\smallskip
\noindent\textbf{Impact of Secret Tokens.}
We first examine the contribution of secret tokens (STs) in Table \ref{tab:ablation-study} (Line 1). 
As explained in Appendix \ref{sec:implementation-details}, we do not apply STs to CoLA and report the corresponding setting as N/A. On GSM8K, removing STs slightly increases utility from 0.455 to 0.464, but substantially weakens privacy protection. In particular, the SIP reconstruction score increases from 0.042 to 0.267, while the TAG, LAMP, and BiSR scores increase from 0.303, 0.272, and 0.106 to 0.388, 0.381, and 0.211, respectively. This degradation under SIP indicates that STs effectively disrupt the token-level correspondence between the original input and the transmitted representations, making forward inversion substantially more difficult. Their effect on gradient-based and bidirectional attacks further suggests that token-level obfuscation complements representation-level mixing by reducing the sample-specific information retained throughout both forward and backward propagation.

\smallskip
\noindent\textbf{Impact of Calibration Model.}
We next evaluate the calibration model, which refines the approximated representations. Removing this module causes a minor utility decrease on CoLA, but substantially increases the reconstruction scores of SIP, TAG, LAMP, and BiSR from 0.135, 0.240, 0.251, and 0.092 to 0.594, 0.488, 0.475, and 0.181, respectively (Line 3). The effect is more pronounced on GSM8K: utility decreases from 0.455 to 0.350, while the BiSR score increases from 0.106 to 0.358. These results show that the Calibration Model not only mitigates the approximation error, but also allows the approximated representations to be less aligned with the target representations, thereby indirectly reducing privacy leakage.
We additionally remove the Stage-I pre-training procedure while retaining it during Stage-II fine-tuning (Line 4). This variant remains relatively stable on CoLA, but reduces the GSM8K utility from 0.455 to 0.291. The larger degradation on GSM8K indicates that autoregressive generation is more sensitive to inaccurate intermediate representations and that an adequately initialized calibration model is important for private fine-tuning.

\smallskip
\noindent\textbf{Impact of Adaptive Gradient Perturbation.} 
We evaluate adaptive gradient perturbation (GP), which introduces noise based on the calibration residual. On CoLA, removing GP slightly improves utility from 0.846 to 0.853, but consistently weakens privacy protection, particularly against gradient-based attacks. On GSM8K, a similar outcome is observed, where the reconstruction scores of TAG, LAMP, and BiSR increase from 0.303, 0.272, and 0.106 to 0.390, 0.401, and 0.175, respectively (Line 5). Replacing adaptive GP with non-adaptive perturbation also results in lower utility and weaker privacy preservation, although the degradation is relatively modest (Line 6). These results indicate the effectiveness of the residual-aware perturbation strategy. Interestingly, removing GP alone does not improve utility on GSM8K; instead, it reduces performance. To further investigate this observation, we remove both GP and STs. Under this setting, the utility recovers, but the effectiveness of DRAs increases substantially (Line 7). This result indicates that STs and GP form a complementary mechanism. Specifically, STs introduce additional gradient variation, which improves privacy protection but may also increase calibration residual and destabilize optimization. Residual-aware GP adjusts residual-sensitive gradient updates and alleviates this instability.


\smallskip
\noindent\textbf{Impact of Source Mixing Size $k$.}
We investigate the effect of the source mixing size $k$, which determines the number of representations mixed into each obfuscated message. We vary $k$ from 3 to 10 while fixing the message size to $m=3$, and report the results in Figure \ref{fig:varying-k} and Table \ref{tab:all-ablation-study}. 

As shown in Figure \ref{fig:varying-k}(a), increasing $k$ generally improves privacy protection. On CoLA, the RF scores of TAG and LAMP decrease from 0.255 and 0.252 at ($k = 3$) to 0.210 and 0.209 at ($k = 10$), respectively, while the BiSR RF score drops from 0.103 to 0.055. \sys performs similarly on GSM8K. The underlying reason is that mixing a larger number of representations further dilutes the target-specific signals among the support representations, which makes reconstruction more difficult. Figure \ref{fig:varying-k}(b) reveals a task-dependent utility trade-off. On CoLA, the ACC remains stable within a narrow range from 0.836 to 0.852, indicating classification tasks are relatively robust to stronger mixing. In contrast, the EMA on GSM8K decreases from 0.443 to 0.407 as $k$ increases from 3 to 10. This result suggests that multi-step generation is more sensitive to representation approximation errors. As shown in Figure \ref{fig:varying-k}(c), increasing $k$ does not introduce additional communication overhead, which remains constant at 1001 GB on CoLA and 875 GB on GSM8K over the entire training process. However, a larger $k$ increases computation because the client is required to encode and mix more support samples. As $k$ increases from 3 to 10, the training time rises by 41.1\% on CoLA (2889s vs. 4076s) and 46.3\% on GSM8K (2479s vs. 3628s).

As a result, a moderate $k$ offers a balance among privacy protection, model utility, and computation efficiency. Larger values can be adopted when stronger obfuscation is required.

\smallskip
\noindent\textbf{Impact of Dataset Selection.}
We evaluate \sys under different combinations of public and support datasets to examine their sensitivity to auxiliary-data selection. We use Yelp, a classification dataset \cite{zhang2015character}; WikiText, a general-purpose text corpus \cite{merity2016pointer}; and Multi-Source, a corpus comprising samples from multiple tasks \cite{mai2023split}. Figure \ref{fig:dataset} and Table \ref{tab:all-ablation-study} summarize the utility and reconstruction results across all combinations. In general, \sys exhibits limited sensitivity to the choice of auxiliary data. On CoLA, the clean accuracy varies by only 0.0163 across all settings. Similarly, the SIP reconstruction RF scores remain remarkably consistent on both CoLA and GSM8K. Auxiliary-data selection has a slightly larger effect on gradient-based attacks TAG and LAMP, whose RF scores vary by 0.083 and 0.081 on GSM8K, respectively. Nevertheless, the BiSR scores remain consistently low across all configurations. Accordingly, \sys derives its utility preservation and resistance to DRAs primarily from its obfuscation mechanisms rather than from any particular auxiliary corpus. 

Moreover, \sys does not require the public or support datasets to closely match the private downstream data to maintain strong utility. For instance, on the GSM8K generation task, pre-training the calibration model on Yelp and using Multi-Source as the support dataset yields the highest utility among the evaluated configurations. As a result, \sys is effective across diverse choices of auxiliary data, which provides flexibility for practical deployment.

\section{Discussion}
\subsection{Adaptive Attacks}
We further evaluate \sys against the adaptive attack, where the attacker is assumed to know the high-level defense mechanism and deliberately constructs an inverse model to reconstruct the protected target representation. Specifically, the server-side attacker collects obfuscated messages $\mathbf{H}^m$ and the corresponding target representations $\mathbf{h}^{*}$ transmitted from the client to construct training samples for the inverse model $g_{\text{inv}}$. The inverse model is trained by minimizing:
\begin{equation}
    \mathcal{L}_{\text{inv}} = \left\|g_{\text{inv}}\left(\mathbf{H}^{m}\right) - \mathbf{h}^{*}\right\|_{2}^{2}.
\end{equation}

We consider \textbf{Online} and \textbf{Offline} adaptive attack variants based on how the training samples for $g_{\text{inv}}$ are obtained.
For each variant, we further examine two message-processing strategies: Joint and Cancel. Under the \textbf{Joint} strategy, all transmitted messages are concatenated and jointly processed as the input to $g_{\text{inv}}$. This strategy assesses whether the inverse model can learn correlations across the observable messages. Under the \textbf{Cancel} strategy, the attacker exploits the column-wise sum constraint introduced in Section \ref{sec:mixup-based-sl-protocol} and aggregates the messages before inversion. This represents a stronger attack that exploits the client-side cancellation operation. 

\begin{table}[t]
\centering
\scriptsize
\caption{Results under different adaptive attack variants and message-processing strategies. STs denote Secret Tokens.}
\setlength{\tabcolsep}{8pt}
\renewcommand{\arraystretch}{1}
\begin{tabular}{c c | cc cc}
\toprule
\multirow{2}{*}{Variant} &
\multirow{2}{*}{Strategy} &
\multicolumn{2}{c}{CoLA} &
\multicolumn{2}{c}{GSM8K} \\
\cmidrule(lr){3-4} \cmidrule(lr){5-6}
& & RF & ME & RF & ME \\
\midrule
\multirow{2}{*}{Online}
& \multirow{1}{*}{Joint}
& 0.0938 & 0.0782 & 0.1467 & 0.0849 \\

\cmidrule(lr){2-6}
& \multirow{1}{*}{Cancel}
 & 0.0833 & 0.0515 & 0.1060 & 0.0565 \\

\midrule
\multirow{2}{*}{\makecell{Offline \\ (w/ STs)}}
& \multirow{1}{*}{Joint}
& 0.0229 & 0.0142 & 0.0320 & 0.0172 \\

\cmidrule(lr){2-6}
& \multirow{1}{*}{Cancel}
& 0.0277 & 0.0315 & 0.0247 & 0.0176 \\

\midrule
\multirow{2}{*}{\makecell{Offline \\ (w/o STs)}}
& \multirow{1}{*}{Joint}
& 0.1725 & 0.1706 & 0.2711 & 0.2475 \\

\cmidrule(lr){2-6}
& \multirow{1}{*}{Cancel}
& 0.2608 & 0.2940 & 0.2923 & 0.2850 \\

\bottomrule
\end{tabular}
\label{tab:adaptive-attack-results}
\end{table}

\smallskip
\noindent\textbf{Online Adaptive Attack.}
In this variant, the attacker trains the inverse model online during Stage I alongside calibration-model pre-training, and subsequently uses the trained inverse model to reconstruct private data during privacy-preserving fine-tuning in Stage II. This variant gives the attacker access to the true mixed messages and the corresponding target representations, which serve as the inputs and supervision signals for inverse-model training, respectively. As shown in Table \ref{tab:adaptive-attack-results}, the attack remains ineffective on both CoLA and GSM8K, with all RF and ME scores below $0.2$. The Joint strategy generally outperforms the Cancel strategy, while its strongest performance is still modest, reaching only 0.1467 RF and 0.0849 ME on GSM8K. One possible reason is that inverse-model training is restricted to the dataset used for calibration-model pre-training, which may be insufficient for learning an accurate reconstruction function. To explore whether these constraints weaken the attack, we further examine the Offline variant.

\smallskip
\noindent\textbf{Offline Adaptive Attack.} 
To evaluate a stronger adaptive adversary, we allow the attacker to train the inverse model offline using another auxiliary corpus, WikiText, and a larger number (2$\times$) of training epochs. Since the attacker does not have access to the random matrices used for representation-level obfuscation, these matrices are independently simulated during inverse-model training. We additionally evaluate two token-level obfuscation settings. In the \textbf{With Tokens} setting, the attacker randomly samples secret tokens and inserts them into the auxiliary inputs during training. In the \textbf{Without Tokens} setting, no secret tokens are inserted. Table \ref{tab:adaptive-attack-results} summarizes the results under both settings. When randomly sampled secret tokens are used, all reconstruction scores remain below 0.05. This is because the attacker-generated secret tokens differ substantially from those used during private fine-tuning, which introduces a mismatch and reduces the effectiveness of the inverse model. Removing the secret tokens increases the reconstruction scores across datasets, message-processing strategies, and inference settings. Nevertheless, all RF and ME scores remain below 0.30. This limited DRA performance is because the independently simulated obfuscation and approximation fail to reproduce the sample-specific transformations applied during private fine-tuning. As a result, \sys remains effective even when additional data and computational resources are allocated.

\subsection{Computational and Communication Overhead}
\begin{table}[t]
\centering
\scriptsize
\setlength{\tabcolsep}{8pt}
\caption{Comparison of computation and communication cost per sample on CoLA. $*$ denotes estimated values.}
\label{tab:overhead-comparison}
\begin{tabular}{lcccc}
\toprule
\textbf{Method} 
& \multicolumn{1}{c}{\textbf{Computation (s)}} 
& \multicolumn{1}{c}{\textbf{Communication (MB)}} \\
\midrule
No Defense & 0.037   & 8 \\
DP-Forward & 0.038  & 8 \\
EFA        & 4454.9$^{*}$   & 1920$^{*}$ \\
\sys   & 0.113   & 40 \\
\bottomrule
\end{tabular}
\end{table}
We analyze the computational and communication overhead of different SL methods on CoLA and  Qwen3-1.7B, as reported in Table \ref{tab:overhead-comparison}. The No Defense setting serves as the standard SL baseline, which requires 0.037 seconds of computation and 8 MB of client–server communication per sample. DP-Forward retains the baseline communication cost of 8 MB, but introduces additional perturbation operations, increasing the computation time to 0.038 seconds per sample. The FHE-based method, Encryption-friendly Architecture (EFA)\footnote{We use EFA as the FHE-based baseline because its original paper reports detailed empirical measurements of encrypted training overhead.
}, incurs substantially higher costs due to its expensive encrypted operations and ciphertext expansion, with an estimated computation time of 4454.9 seconds and communication cost of 1920 MB per sample. \sys requires 0.113 seconds of computation and 40 MB of communication per sample. This increase relative to the baseline arises from local mixing, blinding, decoding, and calibration, as well as the transmission of multiple obfuscated messages. Nevertheless, \sys remains substantially more efficient than EFA. Its computation and communication costs overhead is within the same order of magnitude as the standard SL baseline while providing substantially stronger privacy protection.

\section{Conclusion}
In this paper, we presented \sys, a mixup-based privacy-preserving split learning framework for LLMs that protects private training data while retaining the downstream utility. \sys mitigates the risk of exposing private samples to the server through token-level obfuscation, representation-level obfuscation, and adaptive gradient perturbation mechanisms. Extensive experiments on classification and generation tasks demonstrate that \sys consistently achieves a strong privacy–utility trade-off against forward-pass, gradient-based, and bidirectional data reconstruction attacks, while remaining robust under adaptive attacks. Additional evaluations show that \sys maintains stable performance across different auxiliary datasets, model scales, architectures, and fine-tuning strategies. These results establish mixup-based obfuscation as a practical and effective direction for privacy-preserving split learning with Large Language Models.

\section*{Ethics Considerations}

This work studies privacy-preserving split learning for large language models. The primary goal of the proposed method is to reduce the risk that a server can reconstruct a client's private training data from intermediate representations or gradients. 

\textbf{Human subjects and datasets.}
This work does not involve recruiting human participants, conducting user studies, or collecting new data from individuals. Our experiments are conducted on publicly available benchmark datasets and public or synthetic auxiliary data. We use these datasets only for aggregate model-utility and reconstruction-risk evaluation, and we do not attempt to identify individuals or link examples to real persons. No real client data from sensitive deployment domains such as healthcare, finance, or education is used in our experiments.

\textbf{Controlled offline evaluation.}
All experiments in this work are conducted in controlled offline environments. We simulate the split-learning setting using locally partitioned models and publicly available benchmark datasets.  
We also do not scan, probe, or attack any external system. The reconstruction attacks considered in this paper are used only as offline evaluation tools to measure privacy leakage under the threat model studied in this work. The work does not identify a specific vulnerability in a deployed system, and therefore does not require coordinated vulnerability disclosure.

\textbf{Dual-use considerations.}
The paper evaluates several data reconstruction attacks in order to measure the privacy leakage of split-learning protocols. Such attacks could be misused by a malicious or honest-but-curious server to infer private inputs or labels from information exchanged during training. We mitigate this risk by using the attacks only as controlled evaluation tools, by reporting aggregate reconstruction metrics rather than sensitive examples. If artifacts are released, they should be accompanied by clear usage restrictions and documentation stating that the attack implementations are intended only for authorized research and auditing of privacy-preserving learning systems.

\textbf{Limitations and responsible deployment.}
\sys is designed to reduce reconstruction risk while preserving model utility, but it should not be interpreted as eliminating all privacy risks.
Deployments involving sensitive personal data should combine \sys with secure communication channels, access control, audit logging, data minimization, and task-specific legal and ethical review. Practitioners should also evaluate the defense under deployment-specific threat models before using it in high-stakes settings.

\section*{LLM usage considerations}
This paper studies large language models as the target model family for privacy-preserving split learning.
LLMs were used for editorial purposes in this
manuscript, and all outputs were inspected by the authors
to ensure accuracy and originality.

\bibliographystyle{IEEEtran}
\bibliography{reference}

@article{zhao2023survey,
  title={A survey of large language models},
  author={Zhao, Wayne Xin and Zhou, Kun and Li, Junyi and Tang, Tianyi and Wang, Xiaolei and Hou, Yupeng and Min, Yingqian and Zhang, Beichen and Zhang, Junjie and Dong, Zican and others},
  journal={arXiv preprint arXiv:2303.18223},
  year={2023}
}

@String(ICLR  = {ICLR})

@String(AAAI = {AAAI})

@String(ICLR = {Int. Conf. Learn. Represent.})

@inproceedings{c,
  title={{BACKDOORL: B}ackdoor attack against competitive reinforcement learning},
  author={Wang, Lun and Javed, Zaynah and Wu, Xian and Guo, Wenbo and Xing, Xinyu and Song, Dawn},
  booktitle={International Joint Conference on Artificial Intelligence},
  year={2021}
}

@inproceedings{vaswani2017attention,
  title={Attention is all you need},
  author={Vaswani, Ashish and Shazeer, Noam and Parmar, Niki and Uszkoreit, Jakob and Jones, Llion and Gomez, Aidan N and Kaiser, {\L}ukasz and Polosukhin, Illia},
  booktitle={Advances in Neural Information Processing Systems},
  pages={5998--6008},
  year={2017}
}

@article{zhang2017mixup,
  title={mixup: Beyond empirical risk minimization},
  author={Zhang, Hongyi and Cisse, Moustapha and Dauphin, Yann N and Lopez-Paz, David},
  journal={arXiv preprint arXiv:1710.09412},
  year={2017}
}

@inproceedings{socher2013recursive,
  title={Recursive deep models for semantic compositionality over a sentiment treebank},
  author={Socher, Richard and Perelygin, Alex and Wu, Jean and Chuang, Jason and Manning, Christopher D and Ng, Andrew Y and Potts, Christopher},
  booktitle={Conference on Empirical Methods in Natural Language Processing},
  pages={1631--1642},
  year={2013}
}

@article{wang2018glue,
  title={GLUE: A multi-task benchmark and analysis platform for natural language understanding},
  author={Wang, Alex and Singh, Amanpreet and Michael, Julian and Hill, Felix and Levy, Omer and Bowman, Samuel R},
  journal={arXiv preprint arXiv:1804.07461},
  year={2018}
}

@article{mai2023split,
  title={Split-and-denoise: Protect large language model inference with local differential privacy},
  author={Mai, Peihua and Yan, Ran and Huang, Zhe and Yang, Youjia and Pang, Yan},
  journal={arXiv preprint arXiv:2310.09130},
  year={2023}
}

@inproceedings{liu2025dualguard,
  title={DualGuard: A Parameter Space Transformation Approach for Bidirectional Defense in Split-Based LLM Fine-Tuning},
  author={Liu, Zihan and Wang, Yizhen and Wang, Rui and Wu, Sai},
  booktitle={Proceedings of the 63rd Annual Meeting of the Association for Computational Linguistics (Volume 1: Long Papers)},
  pages={17065--17080},
  year={2025}
}

@inproceedings{du2023dp,
  title={Dp-forward: Fine-tuning and inference on language models with differential privacy in forward pass},
  author={Du, Minxin and Yue, Xiang and Chow, Sherman SM and Wang, Tianhao and Huang, Chenyu and Sun, Huan},
  booktitle={Proceedings of the 2023 ACM SIGSAC Conference on Computer and Communications Security},
  pages={2665--2679},
  year={2023}
}

@article{shen2023split,
  title={A split-and-privatize framework for large language model fine-tuning},
  author={Shen, Xicong and Liu, Yang and Liu, Huiqi and Hong, Jue and Duan, Bing and Huang, Zirui and Mao, Yunlong and Wu, Ye and Wu, Di},
  journal={arXiv preprint arXiv:2312.15603},
  year={2023}
}

@inproceedings{zimerman2024converting,
  title={Converting transformers to polynomial form for secure inference over homomorphic encryption},
  author={Zimerman, Itamar and Baruch, Moran and Drucker, Nir and Ezov, Gilad and Soceanu, Omri and Wolf, Lior},
  booktitle={Forty-first International Conference on Machine Learning},
  year={2024}
}

@inproceedings{nguyen2023split,
  title={Split without a leak: Reducing privacy leakage in split learning},
  author={Nguyen, Khoa and Khan, Tanveer and Michalas, Antonis},
  booktitle={International Conference on Security and Privacy in Communication Systems},
  pages={321--344},
  year={2023},
  organization={Springer}
}

@article{khan2023split,
  title={Split ways: Privacy-preserving training of encrypted data using split learning},
  author={Khan, Tanveer and Nguyen, Khoa and Michalas, Antonis},
  journal={arXiv preprint arXiv:2301.08778},
  year={2023}
}

@article{pereteanu2022split,
  title={Split HE: Fast secure inference combining split learning and homomorphic encryption},
  author={Pereteanu, George-Liviu and Alansary, Amir and Passerat-Palmbach, Jonathan},
  journal={arXiv preprint arXiv:2202.13351},
  year={2022}
}

@inproceedings{vepakomma2020nopeek,
  title={NoPeek: Information leakage reduction to share activations in distributed deep learning},
  author={Vepakomma, Praneeth and Singh, Abhishek and Gupta, Otkrist and Raskar, Ramesh},
  booktitle={2020 International Conference on Data Mining Workshops (ICDMW)},
  pages={933--942},
  year={2020},
  organization={IEEE}
}

@article{chapelle2000vicinal,
  title={Vicinal risk minimization},
  author={Chapelle, Olivier and Weston, Jason and Bottou, L{\'e}on and Vapnik, Vladimir},
  journal={Advances in neural information processing systems},
  volume={13},
  year={2000}
}

@article{hendrycks2016gaussian,
  title={Gaussian error linear units (gelus)},
  author={Hendrycks, Dan and Gimpel, Kevin},
  journal={arXiv preprint arXiv:1606.08415},
  year={2016}
}

@article{hinton2012improving,
  title={Improving neural networks by preventing co-adaptation of feature detectors},
  author={Hinton, Geoffrey E and Srivastava, Nitish and Krizhevsky, Alex and Sutskever, Ilya and Salakhutdinov, Ruslan R},
  journal={arXiv preprint arXiv:1207.0580},
  year={2012}
}

@article{warstadt2019neural,
  title={Neural network acceptability judgments},
  author={Warstadt, Alex and Singh, Amanpreet and Bowman, Samuel R},
  journal={Transactions of the Association for Computational Linguistics},
  volume={7},
  pages={625--641},
  year={2019},
  publisher={MIT Press One Rogers Street, Cambridge, MA 02142-1209, USA journals-info~…}
}

@article{cobbe2021training,
  title={Training verifiers to solve math word problems, 2021},
  author={Cobbe, Karl and Kosaraju, Vineet and Bavarian, Mohammad and Chen, Mark and Jun, Heewoo and Kaiser, Lukasz and Plappert, Matthias and Tworek, Jerry and Hilton, Jacob and Nakano, Reiichiro and others},
  journal={URL https://arxiv. org/abs/2110.14168},
  volume={9},
  year={2021}
}

@misc{chaudhary2023code,
  title={Code alpaca: An instruction-following llama model for code generation},
  author={Chaudhary, Sahil},
  year={2023}
}

@inproceedings{gliwa2019samsum,
  title={SAMSum corpus: A human-annotated dialogue dataset for abstractive summarization},
  author={Gliwa, Bogdan and Mochol, Iwona and Biesek, Maciej and Wawer, Aleksander},
  booktitle={Proceedings of the 2nd Workshop on New Frontiers in Summarization},
  pages={70--79},
  year={2019}
}

@inproceedings{gardent2017webnlg,
  title={The WebNLG challenge: Generating text from RDF data},
  author={Gardent, Claire and Shimorina, Anastasia and Narayan, Shashi and Perez-Beltrachini, Laura},
  booktitle={Proceedings of the 10th international conference on natural language generation},
  pages={124--133},
  year={2017}
}

@article{grattafiori2024llama,
  title={The llama 3 herd of models},
  author={Grattafiori, Aaron and Dubey, Abhimanyu and Jauhri, Abhinav and Pandey, Abhinav and Kadian, Abhishek and Al-Dahle, Ahmad and Letman, Aiesha and Mathur, Akhil and Schelten, Alan and Vaughan, Alex and others},
  journal={arXiv preprint arXiv:2407.21783},
  year={2024}
}

@article{yang2025qwen3,
  title={Qwen3 technical report},
  author={Yang, An and Li, Anfeng and Yang, Baosong and Zhang, Beichen and Hui, Binyuan and Zheng, Bo and Yu, Bowen and Gao, Chang and Huang, Chengen and Lv, Chenxu and others},
  journal={arXiv preprint arXiv:2505.09388},
  year={2025}
}

@article{hu2022lora,
  title={Lora: Low-rank adaptation of large language models.},
  author={Hu, Edward J and Shen, Yelong and Wallis, Phillip and Allen-Zhu, Zeyuan and Li, Yuanzhi and Wang, Shean and Wang, Liang and Chen, Weizhu and others},
  journal={Iclr},
  volume={1},
  number={2},
  pages={3},
  year={2022}
}

@inproceedings{chen2024unveiling,
  title={Unveiling the vulnerability of private fine-tuning in split-based frameworks for large language models: A bidirectionally enhanced attack},
  author={Chen, Guanzhong and Qin, Zhenghan and Yang, Mingxin and Zhou, Yajie and Fan, Tao and Du, Tianyu and Xu, Zenglin},
  booktitle={Proceedings of the 2024 on ACM SIGSAC Conference on Computer and Communications Security},
  pages={2904--2918},
  year={2024}
}

@inproceedings{deng2021tag,
  title={Tag: Gradient attack on transformer-based language models},
  author={Deng, Jieren and Wang, Yijue and Li, Ji and Wang, Chenghong and Shang, Chao and Liu, Hang and Rajasekaran, Sanguthevar and Ding, Caiwen},
  booktitle={Findings of the association for computational linguistics: EMNLP 2021},
  pages={3600--3610},
  year={2021}
}

@article{balunovic2022lamp,
  title={Lamp: Extracting text from gradients with language model priors},
  author={Balunovic, Mislav and Dimitrov, Dimitar and Jovanovi{\'c}, Nikola and Vechev, Martin},
  journal={Advances in Neural Information Processing Systems},
  volume={35},
  pages={7641--7654},
  year={2022}
}

@inproceedings{lin2004rouge,
  title={Rouge: A package for automatic evaluation of summaries},
  author={Lin, Chin-Yew},
  booktitle={Text summarization branches out},
  pages={74--81},
  year={2004}
}

@inproceedings{papineni2002bleu,
  title={Bleu: a method for automatic evaluation of machine translation},
  author={Papineni, Kishore and Roukos, Salim and Ward, Todd and Zhu, Wei-Jing},
  booktitle={Proceedings of the 40th annual meeting of the Association for Computational Linguistics},
  pages={311--318},
  year={2002}
}

@article{ren2020codebleu,
  title={Codebleu: a method for automatic evaluation of code synthesis},
  author={Ren, Shuo and Guo, Daya and Lu, Shuai and Zhou, Long and Liu, Shujie and Tang, Duyu and Sundaresan, Neel and Zhou, Ming and Blanco, Ambrosio and Ma, Shuai},
  journal={arXiv preprint arXiv:2009.10297},
  year={2020}
}

@inproceedings{banerjee2005meteor,
  title={METEOR: An automatic metric for MT evaluation with improved correlation with human judgments},
  author={Banerjee, Satanjeev and Lavie, Alon},
  booktitle={Proceedings of the acl workshop on intrinsic and extrinsic evaluation measures for machine translation and/or summarization},
  pages={65--72},
  year={2005}
}

@article{rho2024encryption,
  title={Encryption-friendly llm architecture},
  author={Rho, Donghwan and Kim, Taeseong and Park, Minje and Kim, Jung Woo and Chae, Hyunsik and Ryu, Ernest K and Cheon, Jung Hee},
  journal={arXiv preprint arXiv:2410.02486},
  year={2024}
}

@article{zhang2024secure,
  title={Secure transformer inference made non-interactive},
  author={Zhang, Jiawen and Yang, Xinpeng and He, Lipeng and Chen, Kejia and Lu, Wen-jie and Wang, Yinghao and Hou, Xiaoyang and Liu, Jian and Ren, Kui and Yang, Xiaohu},
  journal={Cryptology ePrint Archive},
  year={2024}
}

@inproceedings{moon2025thor,
  title={THOR: Secure transformer inference with homomorphic encryption},
  author={Moon, Jungho and Yoo, Dongwoo and Jiang, Xiaoqian and Kim, Miran},
  booktitle={Proceedings of the 2025 ACM SIGSAC Conference on Computer and Communications Security},
  pages={3765--3779},
  year={2025}
}

@inproceedings{zhang2025moai,
  title={MOAI: Module-optimizing architecture for non-interactive secure transformer inference},
  author={Zhang, Linru and Wang, Xiangning and Sim, Jun Jie and Huang, Zhicong and Zhong, Jiahao and Wang, Huaxiong and Duan, Pu and Lam, Kwok-Yan},
  booktitle={The Fourteenth International Conference on Learning Representations},
  year={2025}
}

@inproceedings{abadi2016deep,
  title={Deep learning with differential privacy},
  author={Abadi, Martin and Chu, Andy and Goodfellow, Ian and McMahan, H Brendan and Mironov, Ilya and Talwar, Kunal and Zhang, Li},
  booktitle={Proceedings of the 2016 ACM SIGSAC conference on computer and communications security},
  pages={308--318},
  year={2016}
}

@article{kanpak2024cure,
  title={Cure: Privacy-preserving split learning done right},
  author={Kanpak, Halil Ibrahim and Shabbir, Aqsa and Gen{\c{c}}, Esra and K{\"u}p{\c{c}}{\"u}, Alptekin and Sav, Sinem},
  journal={arXiv preprint arXiv:2407.08977},
  year={2024}
}

@article{yang2022differentially,
  title={Differentially private label protection in split learning},
  author={Yang, Xin and Sun, Jiankai and Yao, Yuanshun and Xie, Junyuan and Wang, Chong},
  journal={arXiv preprint arXiv:2203.02073},
  year={2022}
}

@inproceedings{zaland2025guarding,
  title={Guarding the middle: Protecting intermediate representations in federated split learning},
  author={Zaland, Obaidullah and Mistry, Sajib and Bhuyan, Monowar},
  booktitle={2025 IEEE International Conference on Big Data (BigData)},
  pages={4425--4434},
  year={2025},
  organization={IEEE}
}

@article{hoffmann2022training,
  title={Training compute-optimal large language models},
  author={Hoffmann, Jordan and Borgeaud, Sebastian and Mensch, Arthur and Buchatskaya, Elena and Cai, Trevor and Rutherford, Eliza and Casas, DDL and Hendricks, Lisa Anne and Welbl, Johannes and Clark, Aidan and others},
  journal={arXiv preprint arXiv:2203.15556},
  volume={10},
  year={2022}
}

@inproceedings{mcmahan2017communication,
  title={Communication-efficient learning of deep networks from decentralized data},
  author={McMahan, Brendan and Moore, Eider and Ramage, Daniel and Hampson, Seth and y Arcas, Blaise Aguera},
  booktitle={Artificial intelligence and statistics},
  pages={1273--1282},
  year={2017},
  organization={Pmlr}
}

@inproceedings{thapa2022splitfed,
  title={Splitfed: When federated learning meets split learning},
  author={Thapa, Chandra and Arachchige, Pathum Chamikara Mahawaga and Camtepe, Seyit and Sun, Lichao},
  booktitle={Proceedings of the AAAI conference on artificial intelligence},
  volume={36},
  number={8},
  pages={8485--8493},
  year={2022}
}

@inproceedings{kariyappa2023exploit,
  title={Exploit: Extracting private labels in split learning},
  author={Kariyappa, Sanjay and Qureshi, Moinuddin K},
  booktitle={2023 IEEE conference on secure and trustworthy machine learning (SaTML)},
  pages={165--175},
  year={2023},
  organization={IEEE}
}

@article{gupta2018distributed,
  title={Distributed learning of deep neural network over multiple agents},
  author={Gupta, Otkrist and Raskar, Ramesh},
  journal={Journal of Network and Computer Applications},
  volume={116},
  pages={1--8},
  year={2018},
  publisher={Elsevier}
}

@inproceedings{feng2026mitigating,
  title={Mitigating gradient inversion risks in language models via token obfuscation},
  author={Feng, Xinguo and Ma, Zhongkui and Wang, Zihan and Abuadbba, Alsharif and Bai, Guangdong},
  booktitle={Proceedings of the ACM Asia Conference on Computer and Communications Security},
  pages={1832--1848},
  year={2026}
}

@inproceedings{sun2020mixup,
  title={Mixup-transformer: Dynamic data augmentation for NLP tasks},
  author={Sun, Lichao and Xia, Congying and Yin, Wenpeng and Liang, Tingting and Yu, Philip S and He, Lifang},
  booktitle={Proceedings of the 28th International Conference on Computational Linguistics},
  pages={3436--3440},
  year={2020}
}

@inproceedings{chen2020mixtext,
  title={Mixtext: Linguistically-informed interpolation of hidden space for semi-supervised text classification},
  author={Chen, Jiaao and Yang, Zichao and Yang, Diyi},
  booktitle={Proceedings of the 58th annual meeting of the association for computational linguistics},
  pages={2147--2157},
  year={2020}
}

@article{pham2024enhancing,
  title={Enhancing accuracy-privacy trade-off in differentially private split learning},
  author={Pham, Ngoc Duy and Phan, Khoa T and Chilamkurti, Naveen},
  journal={IEEE Transactions on Emerging Topics in Computational Intelligence},
  volume={9},
  number={1},
  pages={988--1000},
  year={2024},
  publisher={IEEE}
}

@article{bommasani2021opportunities,
  title={On the opportunities and risks of foundation models},
  author={Bommasani, Rishi and Hudson, Drew A and Adeli, Ehsan and Altman, Russ and Arora, Simran and von Arx, Sydney and Bernstein, Michael S and Bohg, Jeannette and Bosselut, Antoine and Brunskill, Emma and others},
  journal={arXiv preprint arXiv:2108.07258},
  year={2021}
}

@article{lin2024split,
  title={Split learning in 6G edge networks},
  author={Lin, Zheng and Qu, Guanqiao and Chen, Xianhao and Huang, Kaibin},
  journal={IEEE Wireless Communications},
  volume={31},
  number={4},
  pages={170--176},
  year={2024},
  publisher={IEEE}
}

@inproceedings{lyu2023optimal,
  title={Optimal resource allocation for u-shaped parallel split learning},
  author={Lyu, Song and Lin, Zheng and Qu, Guanqiao and Chen, Xianhao and Huang, Xiaoxia and Li, Pan},
  booktitle={2023 IEEE Globecom Workshops (GC Wkshps)},
  pages={197--202},
  year={2023},
  organization={IEEE}
}

@inproceedings{pham2023data,
  title={Data leakage threats and protection in split learning: a survey},
  author={Pham, Ngoc Duy and Chilamkurti, Naveen},
  booktitle={Proceedings of the 2023 International Conference on Intelligent Computing and Its Emerging Applications},
  pages={141--147},
  year={2023}
}

@article{zhang2015character,
  title={Character-level convolutional networks for text classification},
  author={Zhang, Xiang and Zhao, Junbo and LeCun, Yann},
  journal={Advances in neural information processing systems},
  volume={28},
  year={2015}
}

@article{merity2016pointer,
  title={Pointer sentinel mixture models},
  author={Merity, Stephen and Xiong, Caiming and Bradbury, James and Socher, Richard},
  journal={arXiv preprint arXiv:1609.07843},
  year={2016}
}

\appendices

\section{Experimental Details}\label{sec:experimental-details}

\subsection{Evaluation Metrics.} For completeness, we provide the formal definitions of all evaluation metrics used in this work.


\textit{Accuracy (ACC).} For CoLA, SST-2, MRPC and RTE, we use classification accuracy. Given $N$ test examples, predicted labels $\hat{y}_i$, and gold labels $y_i$, the accuracy is computed as
\begin{equation}
\text{Accuracy}=\frac{1}{N}\sum_{i=1}^{N}\mathbf{1}(\hat{y}_i=y_i),
\end{equation}
where $\mathbf{1}(\cdot)$ is the indicator function that returns 1 if the condition is true and 0 otherwise.

\textit{Exact Match Accuracy (EMA).} For GSM8K, we use answer accuracy under exact match on the final answer. Let $a_i$ denote the gold final numeric answer for the $i$-th sample, and let $\hat{a}_i$ denote the extracted final answer from the model output. The metric is
\begin{equation}
\text{ACC}_{\text{EM}}=
\frac{1}{N}\sum_{i=1}^{N}\mathbf{1}(\hat{a}_i=a_i).
\end{equation}

\textit{BLEU.} For WebNLG, we use BLEU as the evaluation metric. BLEU is defined as
\begin{equation}
\text{BLEU} = \text{BP}\cdot
\exp\left(
\sum_{n=1}^{N} w_n \log p_n
\right),
\end{equation}
where $p_n$ is the modified $n$-gram precision for $n$-grams of order $n$, $w_n$ is the corresponding weight, and BP is the brevity penalty:
\begin{equation}
\text{BP}=
\begin{cases}
1, & c>r,\\[4pt]
\exp(1-r/c), & c \le r,
\end{cases}  
\end{equation}
with $c$ being the generated length and $r$ the reference length.

\textit{CodeBLEU.} For CodeAlpaca-20K, we use CodeBLEU to evaluate code generation quality. CodeBLEU extends BLEU by additionally incorporating syntax- and semantics-related information. Its general form is
\begin{equation}
\begin{aligned}
\text{CodeBLEU}
&=
\alpha \cdot \text{BLEU}
+
\beta \cdot \text{WNgram}
+
\gamma \cdot \text{SyntaxMatch} \\
&\quad
+
\delta \cdot \text{DataflowMatch}.
\end{aligned}
\end{equation}
where $\alpha$,$\beta$,$\gamma$,$\delta \geq$0 and satisfy
\begin{equation}
\alpha+\beta+\gamma+\delta = 1.
\end{equation}
Here, BLEU denotes the standard n-gram overlap score, WNgram is a weighted $n$-gram match emphasizing keywords and important code tokens, SyntaxMatch measures similarity in abstract syntax tree structures, and DataflowMatch evaluates semantic consistency through data-flow relations.

\textit{ROUGE-L F1 (RF).}
We report ROUGE-L F1 to evaluate summary quality on SAMSum, where a higher score indicates greater similarity to the reference summary. It measures the in-order token overlap between a generated text $\hat{Y}$ and a reference text $Y$ based on their longest common subsequence (LCS).
Let $m$ and $n$ denote the lengths of $Y$ and $\hat{Y}$, respectively. The precision and recall are
\begin{equation}
P_{\mathrm{LCS}} =
\frac{\operatorname{LCS}(Y, \hat{Y})}{n},
\qquad
R_{\mathrm{LCS}} =
\frac{\operatorname{LCS}(Y, \hat{Y})}{m}.
\end{equation}

ROUGE-L F1 is computed as their harmonic mean:
\begin{equation}
\mathrm{RF} =
\frac{2P_{\mathrm{LCS}}R_{\mathrm{LCS}}}
{P_{\mathrm{LCS}}+R_{\mathrm{LCS}}}.
\end{equation}

We also use ROUGE-L F1 to evaluate DRAs by comparing the reconstructed text with the original private text. A lower RF score indicates stronger privacy protection.

\textit{METEOR (ME).}
We further report METEOR to evaluate the lexical-semantic similarity between the original private text and the reconstructed text. METEOR aligns unigrams based on exact, stem, and synonym matches. Let $M$ denote the number of aligned tokens, and let $|\hat{Y}|$ and $|Y|$ denote the lengths of the reconstructed and reference texts, respectively. The unigram precision, recall and recall-weighted harmonic mean are computed as
\begin{equation}
P_{\mathrm{ME}}=\frac{M}{|\hat{Y}|}, \;
R_{\mathrm{ME}}=\frac{M}{|Y|}, \;
F_{\mathrm{mean}}=
\frac{10P_{\mathrm{ME}}R_{\mathrm{ME}}}
{R_{\mathrm{ME}}+9P_{\mathrm{ME}}}.
\end{equation}
To penalize fragmented alignments, let $C$ denote the number of contiguous matched chunks. 
The final METEOR score is
\begin{equation}
\mathrm{Penalty} =
\gamma \left(\frac{C}{M}\right)^{\theta},
\qquad
\mathrm{ME} =
(1-\mathrm{Penalty})F_{\mathrm{mean}},
\end{equation}
where $\gamma=0.5$ and $\theta=3$ follow the standard configuration. A lower ME score indicates stronger resistance to data reconstruction attacks.

\subsection{Implementation Details}\label{sec:implementation-details}
 During token-level obfuscation, secret tokens are sampled uniformly at random from the model vocabulary and inserted at random positions in each input sequence. We insert 20 secret tokens for classification tasks and 40 secret tokens for generation tasks. For representation-level obfuscation, we set both the size of input representations $k$ and output messages $m$ to 3. The entries of the secret mixing matrix $\mathbf{A}$ and blinding matrix $\mathbf{B}$ are randomly generated subject to the constraints described in Section \ref{sec:mixup-based-sl-protocol}.

The calibration model consists of one low-rank residual block with a bottleneck dimension of $r=64$ and a dropout rate of $0.1$. During calibration model training, we train the calibration model for two epochs using AdamW with a learning rate of $1 \times 10^{-4}$ and a batch size of $4$. During privacy-preserving fine-tuning, we train each model for $3$ epochs using AdamW with a batch size of $4$ and a learning rate of $2 \times 10^{-5}$. All inputs are padded or truncated to a maximum sequence length of 512 tokens before being passed through the model. We alternate global-model optimization and calibration-model refinement at a $1:1$ ratio: each privacy-preserving fine-tuning update is followed by one calibration-model refinement step using public data. For the public and support dataset, we use the Multi-Source corpus constructed by Mai et al.~\cite{mai2023split}, which contains approximately $100,000$ samples drawn from $20$ heterogeneous public text datasets. The gradient perturbation is based on the zero-mean Gaussian distribution, with its magnitude adaptively determined by the calibration residual. For classification tasks, we set the perturbation coefficient to $\gamma=1$ for Llama3.2-1B and $\gamma=0.1$ for Qwen3-1.7B. For generation tasks, the corresponding values are $\gamma=80$  and $\gamma=100$, respectively. During evaluation, we remove the inserted secret tokens before measuring the effectiveness of the DRAs to ensure a fair comparison with the baselines. 

Since CoLA evaluates grammatical acceptability, inserting random tokens may alter the grammaticality of the original sentence and compromise the validity of its label. Therefore, we do not introduce Secret tokens for this dataset.

\section{More Experiments and Results}
\subsection{Inference-Time Calibration.}
\begin{table}[t]
    \centering
    \caption{Utility performance with and without the calibration model. 
    $\Delta$ denotes the performance change.}
    \label{tab:calibration-effectiveness-during-inference}
    \resizebox{\columnwidth}{!}{
    \begin{tabular}{l|cccc|cccc}
        \toprule
        \multirow{1}{*}{Dataset}
        & \multicolumn{4}{c|}{Qwen3-1.7B} 
        & \multicolumn{4}{c}{Llama3-1B} \\
        \cmidrule(lr){2-5} \cmidrule(lr){6-9} 
        & Base & + Cal. & $\Delta$ & $\Delta\%$
        & Base & + Cal. & $\Delta$ & $\Delta\%$\\
        \midrule
        CoLA             & 0.845 & 0.844 & -0.001 & -0.12\% & 0.839 & 0.837 & -0.002 & -0.23\% \\
        MRPC             & 0.879 & 0.884 & +0.005 & +0.56\% & 0.723 & 0.725 & +0.002 & +0.27\% \\
        \midrule
        GSM8K            & 0.459 & 0.446 & -0.013 & -2.83\%& 0.238 & 0.238 & 0.000 & 0.00\% \\
        SAMSum           & 0.533 & 0.532 & -0.001 & -0.19\%& 0.520 & 0.521 & +0.001 & 0.19\% \\
        \bottomrule
    \end{tabular}
    }
\end{table}
During privacy-preserving fine-tuning, \sys employs a lightweight calibration model to mitigate the approximation errors introduced by representation-level obfuscation. At inference time, however, the client can optionally omit this module to reduce local computation and latency. Table \ref{tab:calibration-effectiveness-during-inference} compares the downstream utility of the trained model with and without inference-time calibration. Across all evaluated datasets and model families, removing the calibration model causes only minor performance variations: the absolute difference remains below 0.02, while the relative change is consistently below 3\%. Therefore, the client can generally remove this module after training to reduce inference-time overhead with negligible utility degradation. 

\subsection{Reproducibility}\label{sec:reproducibility}
\begin{figure}[t]
    \centering
    \includegraphics[width=\columnwidth]{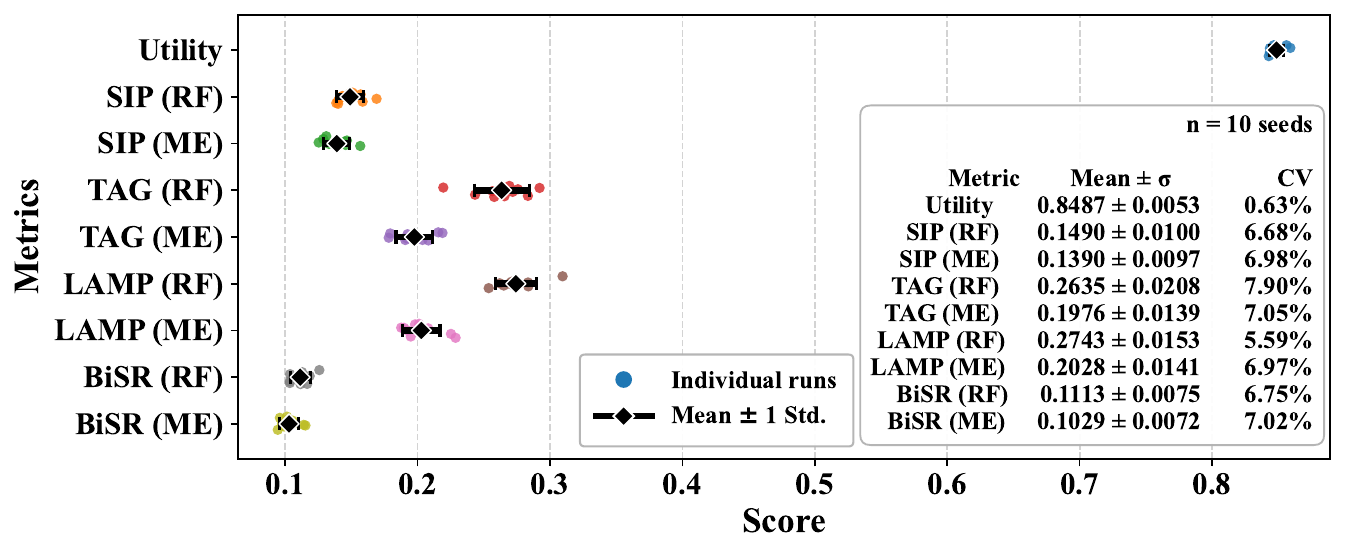}
    \caption{Performance of \sys across 10 independent random seeds.} 
    \label{fig:stability}
\end{figure}
\sys introduces several stochastic operations, including secret token and support-sample selection, random matrix sampling, and gradient perturbation. We therefore evaluate whether its performance is sensitive to random initialization by repeating the full training and evaluation pipeline with 10 independent random seeds. Figure \ref{fig:stability} presents the individual results, together with the mean, standard deviation, and coefficient of variation (CV) for downstream utility and the reconstruction scores on CoLA.

As shown in Figure \ref{fig:stability}, \sys exhibits stable performance across runs. The downstream utility remains consistent at $0.8487 \pm 0.0053$, with a CV of only 0.63\%. The reconstruction scores also show limited variation, with CV values below 8\% across all evaluated metrics, ranging from 5.59\% to 7.90\%. Notably, this consistency is observed across different DRAs. These results demonstrate that \sys is not sensitive to a particular random seed and reliably maintains its performance across independent runs.

\begin{table*}[t]
\centering
\caption{Full results of ablation study on CoLA and GSM8K under different settings.}
\scriptsize
\setlength{\tabcolsep}{2.5pt}
\renewcommand{\arraystretch}{0.9}
\begin{tabular}{l c c c c c c c c c c c c c c c c c c c}
\toprule
\multirow{3}{*}{Setting}
& \multicolumn{9}{c}{CoLA} 
& \multicolumn{9}{c}{GSM8K} \\
\cmidrule(lr){2-10} \cmidrule(lr){11-19}
& Utility & \multicolumn{2}{c}{SIP} & \multicolumn{2}{c}{TAG} & \multicolumn{2}{c}{LAMP} & \multicolumn{2}{c}{BiSR} 
& Utility & \multicolumn{2}{c}{SIP} & \multicolumn{2}{c}{TAG} & \multicolumn{2}{c}{LAMP} & \multicolumn{2}{c}{BiSR} \\
\cmidrule(lr){2-2} \cmidrule(lr){3-4} \cmidrule(lr){5-6} \cmidrule(lr){7-8} \cmidrule(lr){9-10}
\cmidrule(lr){11-11} \cmidrule(lr){12-13} \cmidrule(lr){14-15} \cmidrule(lr){16-17} \cmidrule(lr){18-19}
& ACC & RF & ME & RF & ME & RF & ME & RF & ME & EMA & RF & ME & RF & ME & RF & ME & RF & ME \\
\midrule

\sys & 0.846 & 0.135 & 0.128 & 0.240 & 0.193 & 0.251 & 0.196 & 0.092 & 0.081 & 0.455 & 0.042 & 0.024 & 0.303 & 0.349 & 0.272 & 0.305 & 0.106 & 0.066 \\

\midrule
\multicolumn{19}{l}{\textbf{Secret Tokens (STs)}} \\
\hspace{0.2cm} w/o Secret Tokens & 0.855 & 0.143 & 0.134 & 0.278 & 0.189 & 0.274 & 0.193 & 0.098 & 0.100 & 0.464 & 0.267 & 0.195 & 0.388 & 0.387 & 0.381 & 0.378 & 0.211 & 0.159 \\
\midrule
\multicolumn{19}{l}{\textbf{Calibration Model}} \\
\hspace{0.2cm} w/o Calibration Model & 0.833 & 0.594 & 0.598 & 0.488 & 0.462 & 0.475 & 0.456 & 0.181 & 0.172 & 0.350 & 0.098 & 0.078 & 0.565 & 0.748 & 0.566 & 0.748 & 0.358 & 0.435 \\
\hspace{0.2cm} w/o Pretrain & 0.829 & 0.137 & 0.125 & 0.269 & 0.202 & 0.272 & 0.201 & 0.097 & 0.096 & 0.291 & 0.021 & 0.012 & 0.380 & 0.396 & 0.367 & 0.352 & 0.121 & 0.082 \\
\midrule
\multicolumn{19}{l}{\textbf{Gradient Perturbation}} \\
\hspace{0.2cm} w/o Gradient Perturbation & 0.853 & 0.140 & 0.133 & 0.268 & 0.198 & 0.257 & 0.195 & 0.109 & 0.104 & 0.309 & 0.043 & 0.026 & 0.390 & 0.374 & 0.401 & 0.389 & 0.175 & 0.146 \\
\hspace{0.2cm} w/ non-adaptive Gradient Perturbation & 0.852 & 0.141 & 0.126 & 0.280 & 0.208 & 0.278 & 0.197 & 0.103 & 0.100 & 0.306 & 0.042 & 0.019 & 0.370 & 0.357 & 0.376 & 0.362 & 0.157 & 0.126 \\
\hspace{0.2cm} w/o Secret Tokens and Perturbation & 0.859 & 0.139 & 0.133 & 0.274 & 0.210 & 0.269 & 0.205 & 0.107 & 0.102 & 0.479 & 0.266 & 0.195 & 0.475 & 0.423 & 0.468 & 0.410 & 0.255 & 0.190 \\

\midrule
\multicolumn{19}{l}{\textbf{Source Mixing Size $k$}} \\
\hspace{0.2cm} m=3, k=3  & 0.845 & 0.139 & 0.129 & 0.255 & 0.202 & 0.252 & 0.195 & 0.103 & 0.091 & 0.443 & 0.043 & 0.024 & 0.298 & 0.345 & 0.266 & 0.296 & 0.103 & 0.066 \\
\hspace{0.2cm} m=3, k=4  & 0.850 & 0.134 & 0.126 & 0.234 & 0.192 & 0.235 & 0.189 & 0.068 & 0.069 & 0.438 & 0.043 & 0.022 & 0.322 & 0.349 & 0.294 & 0.309 & 0.104 & 0.067 \\
\hspace{0.2cm} m=3, k=5  & 0.852 & 0.137 & 0.128 & 0.225 & 0.183 & 0.227 & 0.191 & 0.071 & 0.070 & 0.434 & 0.038 & 0.021 & 0.299 & 0.337 & 0.271 & 0.289 & 0.099 & 0.066 \\
\hspace{0.2cm} m=3, k=6  & 0.848 & 0.138 & 0.125 & 0.241 & 0.206 & 0.218 & 0.179 & 0.067 & 0.064 & 0.441 & 0.038 & 0.020 & 0.263 & 0.299 & 0.243 & 0.265 & 0.085 & 0.057 \\
\hspace{0.2cm} m=3, k=7  & 0.850 & 0.137 & 0.125 & 0.221 & 0.193 & 0.213 & 0.167 & 0.063 & 0.065 & 0.426 & 0.034 & 0.017 & 0.291 & 0.315 & 0.255 & 0.275 & 0.090 & 0.055 \\
\hspace{0.2cm} m=3, k=8  & 0.838 & 0.135 & 0.123 & 0.222 & 0.188 & 0.211 & 0.178 & 0.065 & 0.066 & 0.416 & 0.045 & 0.024 & 0.272 & 0.290 & 0.236 & 0.247 & 0.090 & 0.053 \\
\hspace{0.2cm} m=3, k=9  & 0.839 & 0.136 & 0.123 & 0.214 & 0.180 & 0.213 & 0.179 & 0.064 & 0.069 & 0.419 & 0.042 & 0.022 & 0.273 & 0.300 & 0.236 & 0.244 & 0.089 & 0.051 \\
\hspace{0.2cm} m=3, k=10 & 0.836 & 0.131 & 0.119 & 0.210 & 0.178 & 0.209 & 0.178 & 0.055 & 0.065 & 0.407 & 0.038 & 0.018 & 0.271 & 0.301 & 0.231 & 0.248 & 0.084 & 0.047 \\
\midrule
\multicolumn{19}{l}{\textbf{Message Size $m$}} \\
\hspace{0.2cm} m=5, k=3  & 0.856 & 0.094 & 0.092 & 0.203 & 0.145 & 0.211 & 0.157 & 0.091 & 0.092 & 0.438 & 0.022 & 0.010 & 0.176 & 0.187 & 0.173 & 0.176 & 0.046 & 0.025 \\
\hspace{0.2cm} m=7, k=3  & 0.853 & 0.093 & 0.092 & 0.177 & 0.125 & 0.191 & 0.138 & 0.072 & 0.081 & 0.442 & 0.024 & 0.010 & 0.191 & 0.187 & 0.186 & 0.181 & 0.052 & 0.028 \\
\hspace{0.2cm} m=9, k=3  & 0.853 & 0.094 & 0.091 & 0.158 & 0.121 & 0.192 & 0.142 & 0.065 & 0.069 & 0.450 & 0.023 & 0.010 & 0.219 & 0.202 & 0.208 & 0.180 & 0.056 & 0.028 \\

\midrule
\multicolumn{19}{l}{\textbf{Public and Support Dataset}} \\
\hspace{0.2cm} Pub: Wiki / Sup: Wiki & 0.851 & 0.143 & 0.135 & 0.278 & 0.225 & 0.264 & 0.213 & 0.099 & 0.102 & 0.428 & 0.029 & 0.014 & 0.300 & 0.342 & 0.277 & 0.299 & 0.082 & 0.049 \\
\hspace{0.2cm} Pub: Wiki / Sup: Multi-Source & 0.841 & 0.143 & 0.134 & 0.262 & 0.198 & 0.269 & 0.202 & 0.095 & 0.090 & 0.447 & 0.030 & 0.014 & 0.293 & 0.328 & 0.254 & 0.277 & 0.075 & 0.047 \\
\hspace{0.2cm} Pub: Wiki / Sup: Yelp & 0.844 & 0.143 & 0.135 & 0.244 & 0.194 & 0.249 & 0.193 & 0.086 & 0.093 & 0.447 & 0.030 & 0.014 & 0.257 & 0.311 & 0.228 & 0.261 & 0.072 & 0.047 \\
\hspace{0.2cm} Pub: Multi-Source / Sup: Wiki & 0.850 & 0.143 & 0.134 & 0.275 & 0.208 & 0.256 & 0.200 & 0.091 & 0.095 & 0.403 & 0.029 & 0.014 & 0.281 & 0.307 & 0.257 & 0.272 & 0.079 & 0.049 \\
\hspace{0.2cm} Pub: Multi-Source / Sup: Multi-Source & 0.855 & 0.144 & 0.135 & 0.261 & 0.191 & 0.252 & 0.190 & 0.102 & 0.104 & 0.431 & 0.029 & 0.014 & 0.341 & 0.373 & 0.309 & 0.327 & 0.082 & 0.051 \\
\hspace{0.2cm} Pub: Multi-Source / Sup: Yelp & 0.846 & 0.143 & 0.134 & 0.233 & 0.187 & 0.240 & 0.187 & 0.088 & 0.093 & 0.409 & 0.029 & 0.014 & 0.276 & 0.298 & 0.256 & 0.265 & 0.078 & 0.049 \\
\hspace{0.2cm} Pub: Yelp / Sup: Wiki & 0.853 & 0.143 & 0.135 & 0.272 & 0.209 & 0.262 & 0.209 & 0.095 & 0.093 & 0.406 & 0.029 & 0.013 & 0.304 & 0.335 & 0.279 & 0.301 & 0.078 & 0.049 \\
\hspace{0.2cm} Pub: Yelp / Sup: Multi-Source & 0.857 & 0.143 & 0.134 & 0.259 & 0.209 & 0.268 & 0.210 & 0.103 & 0.107 & 0.456 & 0.029 & 0.014 & 0.301 & 0.340 & 0.263 & 0.284 & 0.073 & 0.049 \\
\hspace{0.2cm} Pub: Yelp / Sup: Yelp & 0.851 & 0.143 & 0.135 & 0.255 & 0.200 & 0.236 & 0.190 & 0.080 & 0.084 & 0.420 & 0.030 & 0.014 & 0.278 & 0.324 & 0.254 & 0.281 & 0.075 & 0.050 \\
\bottomrule
\end{tabular}
\label{tab:all-ablation-study}
\end{table*}

\subsection{Impact of source size $k$ and message size $m$.}
In Section \ref{sec:ablation-study} and Figure \ref{fig:varying-k}, we analyze the effect of the source size $k$. Table \ref{tab:all-ablation-study} provides the complete results. 

We further investigate the message size $m$, which controls the number of obfuscated messages transmitted to the server. When fixing $k=3$, increasing $m$ from 3 to 5 can improve resistance to DRAs while preserving downstream utility. For example, on GSM8K, the RF scores of TAG, LAMP, and BiSR decrease from 0.298, 0.266, and 0.103 to 0.176, 0.173, and 0.046, respectively. On CoLA, the SIP RF score also decreases from 0.139 to 0.094, while the clean accuracy slightly increases from 0.845 to 0.856. However, further increasing $m$ from 5 to 9 provides marginal benefits. On CoLA, some reconstruction scores continue to decrease, but the improvement becomes progressively smaller. On GSM8K, reconstruction scores increase slightly when $m$ exceeds 5. For instance, the TAG RF score increases from 0.176 at $m=5$ to 0.219 at $m=9$, while the LAMP RF score increases from 0.173 to 0.208. Meanwhile, both computational and communication overhead increase as $m$ grows. Overall, these results show that simply transmitting more obfuscated messages does not consistently improve the privacy--utility trade-off. A moderate configuration of $m$ provides a more practical balance among privacy protection, model utility, and efficiency.

\subsection{Overhead Estimation of FHE-based Method.}
Direct FHE-based fine-tuning of Qwen3-1.7B is computationally prohibitive. We therefore estimate its overhead using the runtime measurements of Rho et al. \cite{rho2024encryption} and the ciphertext-packing configurations adopted by recent FHE-based Transformer systems \cite{zhang2024secure,moon2025thor,zhang2025moai}. 

\smallskip
\noindent\textbf{Computation Overhead.} Rho et al. \cite{rho2024encryption} report that FHE-based full fine-tuning of a two-layer BERT-style Transformer on CoLA requires 67.1 hours per epoch using eight NVIDIA GeForce RTX 4090 GPUs. With 8,551 training samples, this corresponds to approximately 28.2 seconds per sample per epoch, or 14.1 seconds per sample per layer.

To extrapolate this result to Qwen3-1.7B, we use the number of parameters per Transformer layer as an approximate scaling factor. A BERT-style layer contains about 7 million parameters, whereas a layer in Qwen3-1.7B contains approximately 50 million parameters, which is roughly $7.1\times$ larger. Assuming encrypted fine-tuning cost scales linearly with the number of processed parameters, the cost of one Qwen3-1.7B layer is estimated as $14.1\text{s} \times 7.1 \approx 100.7 \text{s}.$ Since our U-shaped split-learning configuration evaluates 28 intermediate layers, the estimated fine-tuning cost is therefore $100.7\text{s} \times 28 \approx 2819.6 \text{s}$ per sample. According to NVIDIA's official specifications, the aggregate peak BF16 Tensor Core throughput of eight GeForce RTX 4090 GPUs is approximately $1.58\times$ that of a single H200 NVL GPU (2,643 vs. 1,671 TFLOPS). The estimated cost on one H200 NVL is $2819.6\text{s} \times 1.58 \approx 4454.9 \text{s}$ per sample. This estimate highlights the substantial computational overhead introduced by encrypted training, even under an optimistic linear-scaling approximation. In practice, the actual overhead may be higher because encrypted execution also depends on operation types, ciphertext bootstrapping, and the approximation of non-polynomial functions.

\smallskip
\noindent\textbf{Communication Cost.}
In the U-shaped split-learning configuration, each training step involves 4 communication rounds:
\begin{itemize}[leftmargin=*]
\item The encrypted hidden representation from the client to the server during the forward pass;
\item The encrypted output of the trunk model from the server to the client during the forward pass; 
\item The encrypted gradient from the client to the server-side trunk model during backpropagation;
\item The encrypted gradient from the server to the client-side head model during backpropagation.
\end{itemize}

For an input sequence of length $S = 512$ and hidden dimension $d = 2048$, each transmitted tensor contains approximately $S \times d = 1{,}048{,}576$ numerical values. Recent FHE-based Transformer systems commonly adopt SIMD-style ciphertext packing and 32,768 slots per ciphertext \cite{rho2024encryption,zhang2024secure,moon2025thor,zhang2025moai}, the transmitted tensor requires 32 ciphertext. Assuming each ciphertext is about 15 MB, the online communication overhead is $4 \times 32 \times 15 \text{ MB} = 1920 \text{ MB}$ per sample per training step.

This estimate excludes one-time setup costs, such as key generation and key exchange, as well as implementation-dependent metadata and auxiliary ciphertexts. Consequently, the reported value should be interpreted as an approximate lower-order estimate of the online communication overhead.

\end{document}